\documentclass[lettersize,journal]{IEEEtran}
\usepackage{amsmath,amsfonts}
\usepackage{array}
\usepackage[caption=false,font=normalsize,labelfont=sf,textfont=sf]{subfig}
\usepackage{textcomp}
\usepackage{stfloats}
\usepackage{url}
\usepackage{verbatim}
\usepackage{graphicx}
\usepackage{cite}
\usepackage{graphicx}
\usepackage{amssymb}
\usepackage[colorlinks]{hyperref}
\usepackage{booktabs}
\usepackage{multirow}
\usepackage{pifont}
\usepackage{microtype}

\usepackage{algorithmicx}
\usepackage[ruled,vlined,linesnumbered]{algorithm2e}
\newcommand{\cmark}{\ding{51}}

\def\eg{\emph{e.g.,\ }}

\usepackage{amsmath}
\usepackage{bm}

\begin{document}

\title{Uncertainty-Aware Testing-Time Optimization for 3D Human Pose Estimation}


\author{
Ti Wang, Mengyuan Liu$^{\dagger}$, Hong Liu, Bin Ren, Yingxuan You, Wenhao Li, Nicu Sebe, Xia Li
\thanks{$^{\dagger}$ Corresponding author.}

\thanks{
T. Wang is with the State Key Laboratory of General Artificial Intelligence, Peking University, Shenzhen Graduate School, China, and also with the School of Engineering, École Polytechnique Fédérale de Lausanne, Switzerland (e-mail: ti.wang@epfl.ch).

M. Liu and H. Liu are with the State Key Laboratory of General Artificial Intelligence, Peking University, Shenzhen Graduate School, China. (e-mail: \{liumengyuan, hongliu\}@pku.edu.cn.)

Y. You is with the School of Computer and Communication Sciences, École Polytechnique Fédérale de Lausanne, Switzerland (e-mail: yingxuan.you@epfl.ch).
 
W. Li is with the School of Computer Science and Engineering, Nanyang Technological University, Singapore. (e-mail: wenhao.li@ntu.edu.sg.)

X. Xia is with the Department of Computer Science, ETH Zurich, Zurich 8092, Switzerland (e-mail: xiaxiali@ethz.ch).

B. Ren, is with both the Department of Information, University of Pisa, Italy, and the Department of Information Engineering and Computer Science, University of Trento, Italy. e-mail: bin.ren@unitn.it. 

N. Sebe is with the Department of Information Engineering and Computer Science, University of Trento, Italy (e-mail: niculae.sebe@unitn.it).}
\thanks{This work was supported by National Natural Science Foundation of China (No. 62473007), Shenzhen Innovation in Science and Technology Foundation for The Excellent Youth Scholars (No. RCYX20231211090248064).}
}

\markboth{IEEE Transactions on Multimedia}
{Wang \MakeLowercase{\textit{et al.}}: Uncertainty-Aware Testing-Time Optimization for 3D Human Pose Estimation}

\maketitle

\begin{abstract}

Although data-driven methods have achieved success in 3D human pose estimation, they often suffer from domain gaps and exhibit limited generalization. In contrast, optimization-based methods excel in fine-tuning for specific cases but are generally inferior to data-driven methods in overall performance.
We observe that previous optimization-based methods commonly rely on projection constraint, which only ensures alignment in 2D space, potentially leading to the overfitting problem.
To address this, we propose an Uncertainty-Aware testing-time Optimization (UAO) framework, which keeps the prior information of pre-trained model and alleviates the overfitting problem using the uncertainty of joints.
Specifically, during the training phase, we design an effective 2D-to-3D network for estimating the corresponding 3D pose while quantifying the uncertainty of each 3D joint.
For optimization during testing, the proposed optimization framework freezes the pre-trained model and optimizes only a latent state. Projection loss is then employed to ensure the generated poses are well aligned in 2D space for high-quality optimization. Furthermore, we utilize the uncertainty of each joint to determine how much each joint is allowed for optimization.
The effectiveness and superiority of the proposed framework are validated through extensive experiments on challenging datasets: Human3.6M, MPI-INF-3DHP, and 3DPW. Notably, our approach outperforms the previous best result by a large margin of 5.5\% on Human3.6M. Code is available at \href{https://github.com/xiu-cs/UAO-Pose3D}{https://github.com/xiu-cs/UAO-Pose3D}.
\end{abstract}


\section{Introduction}
Monocular 3D human pose estimation aims to estimate 3D body joints from a single image.
This task plays an important role in various applications,
such as computer animation \cite{yoon2021pose}, human-computer interaction \cite{svenstrup2009pose}, and action recognition \cite{liu2017enhanced}. However, due to the depth ambiguity, it remains a challenging task that has drawn extensive attention in recent years.
With the development of deep learning, data-driven methods \cite{martinez2017simple,zhang2022mixste,li2022mhformer,zou2021modulated,cai2019exploiting} have dominated the field of 3D human pose estimation.
During the training process, these methods implicitly acquire knowledge of camera intrinsic parameters and 3D human pose distributions within specific domains. 
However, these data-driven methods struggle with cross-domain generalization and in-the-wild scenarios \cite{gong2021poseaug,gholami2022adaptpose}. 

Historically, traditional solutions usually formulate the 3D human pose estimation task as a pure optimization-based problem \cite{ramakrishna2012reconstructing,simo2012single,wang2014robust}.
These optimization-based methods offer two significant benefits. 
First, they prevent the model from generating unrealistic or out-of-distribution poses, which ultimately boosts the model's robustness. Second, they provide the flexibility of adaptive adjustments tailored for specific samples, thereby enhancing the model's generalization capabilities.
Although traditional methods can alleviate domain gaps by estimating 3D poses on a case-by-case basis, their performance is often compromised due to the limitations of manually crafted feature extractors \cite{liu2015survey}.
In contrast, data-driven methods leverage neural networks to learn feature representations in an automated manner, enabling them to effectively capture the intricacy and variability inherent in the data.
Recently, some human mesh recovery methods \cite{kim2022meta,guan2022out, guan2021bilevel, kolotouros2019learning,joo2021exemplar} have also adopted optimization strategies to refine network parameters based on certain constraints. We observe that the previous optimization-based methods \cite{kolotouros2019learning, zhang2020inference, joo2021exemplar} typically rely only on the projection constraint.
However, due to the inherent depth ambiguity \cite{li2022mhformer}, several potential 3D poses could be projected to the same 2D pose. 
While alignment in 2D space is assured, this imperfect constraint may lead to overfitting during optimization.
Additionally, previous optimization-based methods typically employ pre-trained networks to obtain a good initialization, subsequently optimizing the network parameters. Nevertheless, this approach involves a vast number of parameters to be optimized, which may disrupt the prior information embedded in the pre-trained network.

\begin{figure}[!t]
\centering
\includegraphics[width=1.0\linewidth]{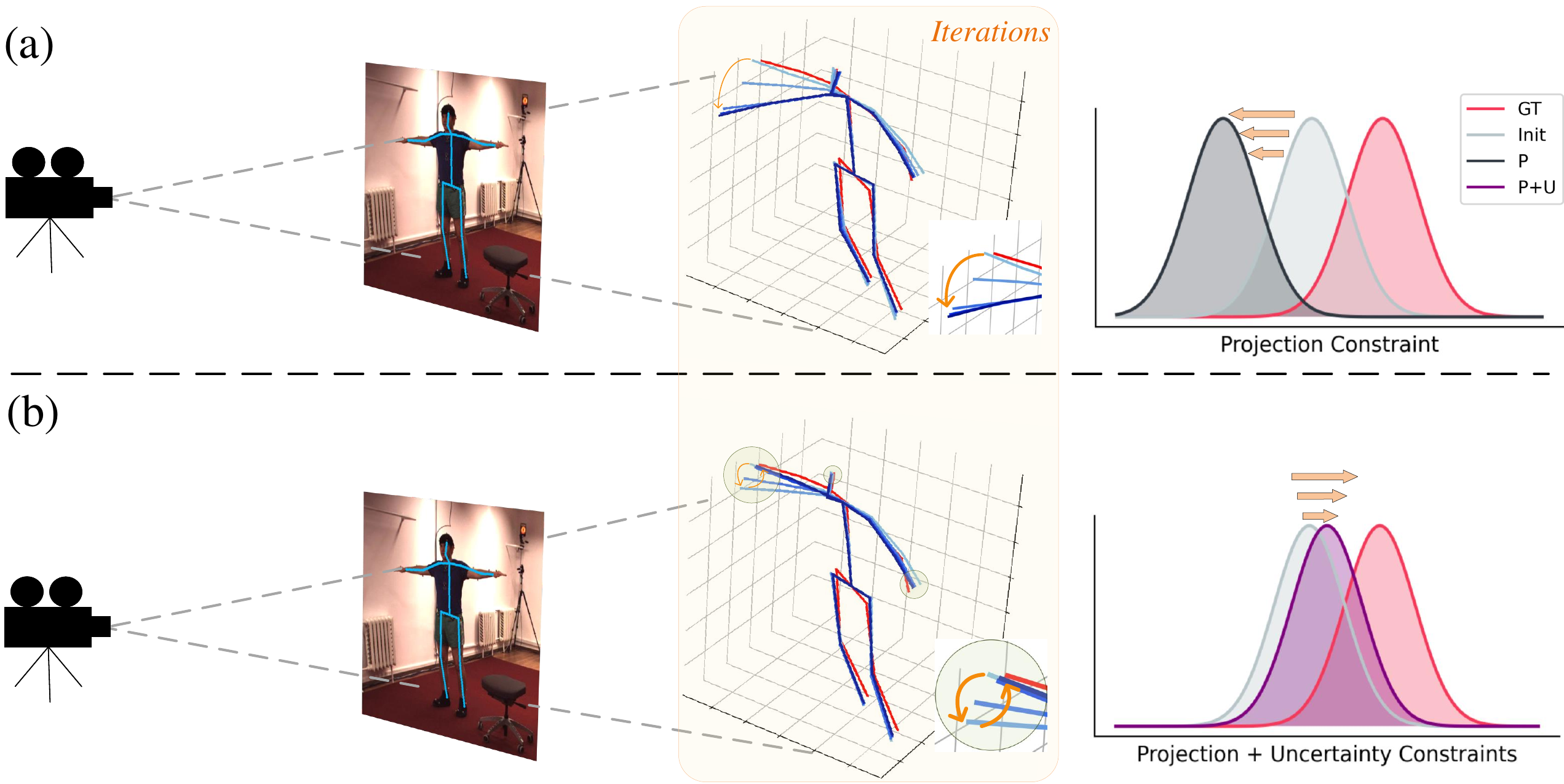}
\caption{
    Schematic diagram of the iterative optimization process. The red pose represents the ground truth, and the blue pose represents the prediction; darker colors indicate later iterations. `GT' refers to the ground truth 3D pose distribution, `Init' denotes the initial result, `P' represents the projection constraint, and `P+U' combines projection and uncertainty constraints.
    (a) Using only the projection constraint during test-time optimization can lead to significant deviations from the ground truth. (b) Incorporating the uncertainty constraint effectively mitigates this issue, yielding final poses that more closely match the ground truth.
}
\label{fig:teaser}
\end{figure}

\textbf{Based on the above observations, we propose an Uncertainty-Aware testing-time Optimization (UAO)} framework for 3D human pose estimation.
This optimization framework retains the prior knowledge embedded in the pre-trained model and alleviates the overfitting problem using the uncertainty of joints.
Specifically, we design an effective network for 2D-to-3D pose lifting, and employ a separate decoder to obtain the uncertainty associated with each joint.  In this way, our network can simultaneously generate the 3D results and the uncertainty of each joint based on input 2D pose during the training phase.
For the testing-time optimization, our target is to make the optimized poses more reliable.
Different from previous optimization-based methods \cite{zhang2020inference, guan2021bilevel, guan2022out, joo2021exemplar, bidulka2025escape, nam2023cyclic} that update the network parameters based on specific constraints during test time, our designed optimization strategy freezes the network parameters and optimizes a latent state.
This design has two benefits: a) fixing the network parameters can keep the learned pose prior during the optimization process; b) setting the optimization target as a hidden state instead of the network parameter only requires the storage of a small number of parameters, making it more suitable for online inference scenarios. Furthermore, our optimization process incorporates a set of well-designed constraints to guide the optimization direction. For 2D pose alignment, we adopt the camera intrinsic parameters to project the generated 3D pose onto the 2D space and minimize the projection loss.
Since we do not know the depth information, achieving precise 3D alignment remains challenging. We observe that relying solely on projection constraint can lead to over-optimization problems, causing the optimized 3D pose to deviate further from the ground truth.
To alleviate this, we introduce an uncertainty constraint that allows 3D joint with higher uncertainty to have a larger divergence radius relative to its initial 3D position, while keeping joint with smaller uncertainty closer to its initial 3D position.
As shown in Figure \ref{fig:teaser}, relying solely on projection constraint may inadvertently steer optimization in the wrong direction, leading to poorer results. This issue is effectively mitigated by further incorporating uncertainty constraints.
With such a unique design, the optimization process is prevented from generating physically implausible poses, leading to more realistic and accurate results.

Our contributions can be summarized as follows:
\begin{itemize}




\item We propose a novel Uncertainty-Aware testing-time Optimization (UAO) framework for 3D human pose estimation. First, an effective 2D-to-3D lifting network is designed to generate 3D outputs while estimating the joint uncertainties. During testing, the pre-trained model is frozen to preserve pose priors, and the uncertainty of each joint is utilized to alleviate the overfitting problem.

\item For test time optimization, a projection constraint is used to ensure that the generated results are aligned in the 2D space. Additionally, we design the uncertainty constraint to determine the degree to which each joint can be optimized, which alleviates the overfitting problem.

\item Extensive experiments demonstrate the effectiveness of the UAO framework and its potential applicability for real-world scenarios.

\end{itemize}

\section{Related Work} \label{section2}
Existing single-view 3D human pose estimation methods fall into two primary categories: a) direct estimation \cite{sun2018integral,ma2021context,pavlakos2017coarse}: infer a 3D human pose from 2D images without the need for intermediate 2D pose representation, b) 2D-to-3D lifting \cite{pavllo20193d,cai2019exploiting,li2025graphmlp,zhang2023learning}: infer a 3D human pose from an intermediately estimated 2D pose. Benefiting from the success of 2D pose detectors \cite{chen2018cascaded,sun2019deep}, the 2D-to-3D lifting approaches generally outperform the direct estimation approaches in both efficiency and effectiveness, becoming the mainstream approach. 
In this work, our approach follows the 2D-to-3D lifting pipeline, which can be further categorized into data-driven and optimization-based methods.

\subsection{Data-driven Methods}
Data-driven approaches involve training a deep neural network to directly regress 3D human poses from provided 2D poses. 
Early attempts simply use fully-connected networks (FCNs) to lift the 2D keypoint to 3D space \cite{martinez2017simple}.
Since the human skeleton can be naturally represented as a graph, various graph convolutional networks (GCNs) based methods \cite {zhao2019semantic,cai2019exploiting,xu2021graph,zou2021modulated} have been proposed.

Recently, transformer-based methods \cite{zheng20213d,strided,li2022mhformer,zhang2022mixste,zhong2023frame,wang2023exploiting,wang2023global} have been widely applied to 3D human pose estimation, achieving new breakthroughs. 
Although these methods can effectively learn pose patterns from datasets, they often exhibit poor generalization ability when faced with unseen data.

\subsection{Optimization-based Methods}
During the early stages, researchers employed pure optimization-based methods \cite{ramakrishna2012reconstructing,simo2012single,wang2014robust} for 3D human pose estimation.
In the era of deep learning, there are also some methods that integrate optimization strategies. We categorize the existing optimization-based methods into two classes.
1) The first class incorporates a regression-based network with optimization strategies, training the entire structure end-to-end, such as SPIN \cite{kolotouros2019learning} and PyMAF \cite{zhang2021pymaf}.
2) The second class follows a post-optimization manner, which is more similar to us, using a pre-trained network to obtain a good initialization, including ISO \cite{zhang2020inference}, BOA \cite{guan2021bilevel}, DynaBOA \cite{guan2022out}, EFT \cite{joo2021exemplar}, ESCAPE \cite{bidulka2025escape} and CycleAdapt \cite{nam2023cyclic}.
However, all these previous post-optimization methods typically optimize the network parameters based on specific constraints, which may disrupt the deep prior knowledge obtained from datasets.

Some approaches also directly optimize the pose and shape parameters of parametric human models (\eg SMPL and SMPL-X), such as SMPLify \cite{bogo2016keep}, SMPLify-X \cite{pavlakos2019expressive}, KBody \cite{zioulis2023kbody}, and HuMoR \cite{rempe2021humor}. But they are constrained by inaccurate initial values or limited representational space.
As a comparison, we set the optimization variable as the predicted 2D pose in latent state, and keep the model parameters frozen during the iterative optimization process.
Besides, pose-related constraints are imperative to ensure the appropriate optimization direction.
Previous optimization-based methods simply rely on the projection constraint \cite{kolotouros2019learning, zhang2020inference, joo2021exemplar}. But due to the depth ambiguity, this imperfect constraint may lead to the wrong optimization direction when over-optimized. To alleviate this problem, we further utilize the uncertainty constraint to ensure that the generated pose lies within the manifold of 3D poses.

\subsection{Uncertainty-based Methods} 
Uncertainty learning has attracted considerable attention and has been applied to numerous tasks, such as object detection \cite{wang2021data}, person re-identification \cite{yu2019robust}, and face recognition \cite{chang2020data,zhang2021relative}.
The key idea is that by knowing the reliability of its outputs, a model can optimize its predictions for improved downstream application performance. In the domain of 2D human pose estimation, confidence scores derived from 2D heatmaps are widely used to assess the reliability of each keypoint \cite{wang2020deep, bogo2016keep}.
For 3D human pose estimation, several studies have explored the use of uncertainty to improve model performance.
Han et al. \cite{han2022single} introduced an uncertainty-based framework for 3D human pose estimation to mitigate the negative impacts of joints with high ambiguous.
Zhang et al. \cite{zhang2022uncertainty} modeled the depth and the 2D distribution uncertainties separately, optimizing these parameters to achieve accurate 3D poses.
In our work, we define keypoint uncertainty as the degree of a keypoint's deviation from its mean position in 3D space, with a higher uncertainty value indicating a larger allowable divergence radius.
Moreover, our approach goes beyond simple uncertainty estimation by incorporating uncertainty constraints into the optimization phase, thereby overcoming the limitations of relying solely on the projection constraint.

\section{Preliminaries} \label{section3}

\subsection{Theory of Uncertainty Estimation}
\label{sec:theory}

Assuming the 3D pose follows Gaussian distributions, we reformulate the 2D-to-3D network as a Bayesian Neural Network (BNN) \cite{jospin2022hands}, which estimates a Gaussian distribution of the target rather than only predicting the absolute coordinates of points.
Specifically, for the 3D joints $\pmb{J}_{i}$ of person $i$ in the dataset, our goal is to estimate the distribution of 3D joints for that individual.
During training, the objective is to maximize the likelihood of all observation pairs.
Let $\theta$ denote the network‘s learnable parameters. The likelihood function $L(\theta)$ can be formulated as follows:
\begin{equation}
    L(\theta) = \prod_{i,k} \mathcal{N}( \pmb{J}_{i,k} | \pmb{\mu}_{i,k}, \Sigma_{i,k}),
\end{equation}
where $k$ is the index for the $k$-th joint of a person, $\pmb{\mu}_{i,k}=f_{\mu}(\theta)$  and $\Sigma_{i,k}=f_{\Sigma}(\theta)$ are the mean and covariance matrix of the Gaussian for the $k$-th joint of person $x_i$. $f_{\mu}$ and $f_{\Sigma}$ are two network heads predicting joint's Gaussian mean and covariance. Each joint is considered as a random variable following a normal Gaussian distribution.
To further elaborate, the mean $\pmb{\mu}_{i,k}$ captures the most probable location of the $k$-th joint, while the covariance matrix $\Sigma_{i,k}$ characterizes the uncertainty of the distribution around this mean. The likelihood term models the probability of observing the actual 3D joint positions given the estimated means and covariances.
In pursuit of the most accurate estimation of the 3D joint positions, we seek the maximum likelihood estimate $\theta^{*}$, which aligns the estimated Gaussian distribution with the observed 3D joint positions, capturing both the most probable locations  $\pmb{\mu}_{i,k}$ and the uncertainty $\Sigma_{i,k}$ associated with the $k$-th joint.
The formulation of the maximum likelihood function can be achieved by:
\begin{align}
    \theta^* &= \arg\max_{\theta} \sum_{i,k}\log \mathcal{N}(\pmb{J}_{i,k} | \pmb{\mu}_{i,k}, \Sigma_{i,k}) \nonumber \\
             &= \arg\min_{\theta}  \frac{1}{2K} \sum_{i,k} \left( (\pmb{J}_{i,k} - \pmb{\mu}_{i,k})^T \Sigma_{i,k}^{-1} (\pmb{J}_{i,k} - \pmb{\mu}_{i,k}) \right. \nonumber \\
             &\left. + \ln|\Sigma_{i,k}| \right),
\end{align}
where $K$ is the total number of joints per pose.
Here we simplify the covariance matrix $\Sigma_{i,k}$ as a diagonal matrix with diagonal elements $\pmb{\sigma}_{i,k}^2$, then we get:
\begin{equation}
    \theta^* = \arg\min_{\theta} \frac{1}{2K} \sum_{i,k}  \left( \frac{||\pmb{J}_{i,k} - \pmb{\mu}_{i,k}||^2}{\pmb{\sigma}_{i,k}^2} + \ln \pmb{\sigma}_{i,k}^2 \right).
\end{equation}
We set $\pmb{s}_{i,k} = \ln \pmb{\sigma}_{i,k}^2$, our target can be reformulated as:
\begin{equation}\label{eq:theta_min_1}
    \theta^* = \arg\min_{\theta} \frac{1}{2K} \sum_{i,k} \left( \frac{ ||\pmb{J}_{i,k} - \pmb{\mu}_{i,k}||^2 }{\exp(\pmb{s}_{i,k})} + \pmb{s}_{i,k} \right).
\end{equation}
Through this formulation, we can minimize the discrepancy between the predicted mean $\pmb{\mu}_i$ and the ground truth $\pmb{J}_{i}$ while accounting for the associated uncertainty $\pmb{s}_i$.
This approach ensures that the estimated Gaussian distribution aligns optimally with the observed 3D joints, providing a robust and reliable characterization of the underlying 3D pose distribution.

\subsection{Graph Convolutional Network}
\label{sec:GCN}
Graph Convolutional Network (GCN) \cite{kipf2016semi} is capable to capture intricate relationships and structures within graph-structured data.
Consider an undirected graph as $G {=} \{V, E\}$, where $V$ is the set of nodes, and $E$ is the set of edges.
The edges can be encoded in an adjacency matrix $A {\in} \{0,1\}^{N {\times} N}$. 
For the input $X_l$ of the $l^{th}$ layer, the vanilla graph convolution aggregates the features of the neighboring nodes. The output  $X_{l+1}$  of the $l^{th}$ GCN layer can be formulated as:
\begin{equation}
    X_{l+1}=\sigma\left(\tilde{D}^{-\frac{1}{2}} \tilde{A} \tilde{D}^{-\frac{1}{2}} X_{l} W \right),
    \label{equ:gcn}
\end{equation}
where $\sigma$ is the ReLU activation function \cite{glorot2011deep}, $W_{l} {\in} \mathbb{R}^{d_{1} \times d} $ is the layer-specific trainable weight matrix.
$\tilde{A} {=} A {+} I_N$ is the adjacency matrix of the graph with added self-connections, where $I_N$ is the identity matrix. Additionally, $\tilde{D}$ is the diagonal node degree matrix.
By stacking multiple GCN layers, it iteratively transforms and aggregates neighboring nodes, thereby obtaining enhanced feature representations.
For the task of 3D human pose estimation, GCN it can effectively capture of the semantic information of the human skeleton.

\section{Uncertainty Estimation}
\label{sec:uncertainty estimation}

For 2D-to-3D lifting, data-driven methods regress the 3D human pose $\hat{\mathbf{J}}^{3D} \in \mathbb{R}^{K\times 3}$ from given 2D pose $\mathbf{J}^{2D}\in \mathbb{R}^{K\times 2}$, 
where $K$ is the number of skeleton joints.
They typically train the model to learn the transformational relationship between 2D pose $\mathbf{J}^{2D}$ and ground truth 3D pose $\mathbf{J}^{3D}$ by minimizing:
\begin{equation}
L_{2D} = ||\mathbf{J}^{3D} - f(\mathbf{J}^{2D})||_{2},
\label{equ:mpjpe}
\end{equation}
where $f$ denotes the 2D-to-3D lifting network for 3D human pose estimation. 
Although this loss function allows the pre-trained model to implicitly encodes inherent patterns for 2D-to-3D lifting and acquire distribution features from the training set, it can not be aware of the uncertainty of each estimated joint.
In order to enable the model to predict the uncertainty of each predicted joint, 
we reformulate the 2D-to-3D network as a Bayesian Neural Network (BNN) \cite{jospin2022hands}, which estimates a Gaussian distribution of the target rather than only predicting the absolute coordinates of points.

Specifically, for the $k$-th joint of the $i$-th person, we let the network predict the expected position $\mathbf{\pmb{\mu}}_{i,k} \in \mathbb{R}^3$ and a simplified scalar covariance $\pmb{\sigma}_{i,k}$ from the input 2D pose $\mathbf{J}^{2D}$. 
By maximizing the likelihood of the training pairs, the loss function to train the BNN over the $i$-th person can be formulated as:
\begin{equation}
\begin{aligned}
    L_{i}^{train} = \frac{1}{2K} \sum_{k=1}^K \left( \frac{\|\mathbf{J}_{i,k}^{3D} - \pmb{\mu}_{i,k}\|^2}{\pmb{\sigma}_{i,k}^2} + \ln \pmb{\sigma}_{i,k}^2 \right) \\
    = \frac{1}{2K} \sum_k^K \left( \|\mathbf{J}_{i,k}^{3D} - \pmb{\mu}_{i,k}\|^2 \exp(-\pmb{s}_{i,k}) + \pmb{s}_{i,k} \right),
\end{aligned}
\end{equation}
where we substitute $\ln \pmb{\sigma}_{i,k}^2$ with $\pmb{s}_{i,k}$ for numerical stability during the training process.

\begin{figure*}[!t]
\centering
\centerline{\includegraphics[width=1\linewidth]{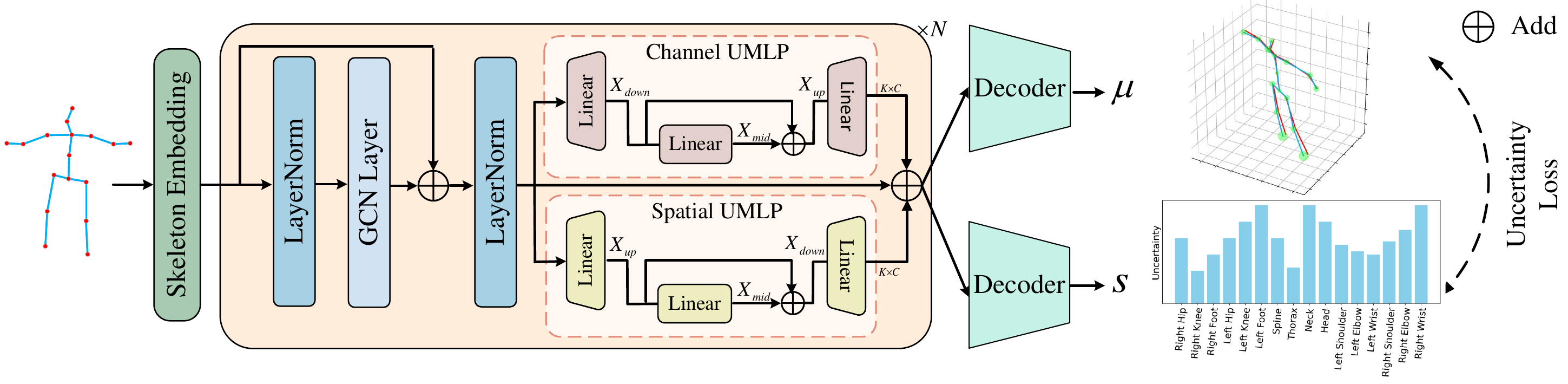}}
\caption
{
Structure of the GUMLP model. Initially, the 2D input is first transformed into high-dimensional features through skeleton embedding. Subsequently, the GCN layer captures the graph information from the high-dimensional features. Then the Channel UMLP and Spatial UMLP are utilized to capture the multi-level and multi-scale information of the skeleton. 
Finally, two decoders are employed to predict the mean pose ${\pmb{\mu}}$ and the uncertainty value $\pmb{s}$ of the corresponding 3D pose distribution.
The model is trained using uncertainty loss, which is calculated based on ${\pmb{\mu}}$ and $\pmb{s}$. 
}
\label{fig:model}
\end{figure*}

To achieve this, we propose GUMLP, primarily consisting of two parts: a GCN layer and a parallel U-shaped multi-layer perception (UMLP) module, as shown in Figure \ref{fig:model}.
In contrast to the conventional data-driven model \cite{martinez2017simple,zou2021modulated}, which typically employs only one decoder to make direct predictions, we utilize two distinct decoders to predict $\pmb{\mu}$ and $\pmb{s}$, respectively.
The GCN layer focuses on local joint relations and introduces the topological priors of the human skeleton. 
For the second component, the UMLP module employs a bottleneck structure \cite{ronneberger2015u, cai2019exploiting} instead of the original MLP in the transformer \cite{Attention}.
The UMLP module operates in two parallel pathways: Channel UMLP and Spatial UMLP, to capture multi-scale and multi-level features effectively.
In the Channel UMLP pathway, the input $\mathbf{X}$ is first processed by a down-projection layer ($\mathbf{X}_{down}$), followed by a middle layer ($\mathbf{X}_{mid}$) that maintains the same dimension, and ultimately, is fed into an up-projection layer ($\mathbf{X}_{up}$).
The simplified formulation can be defined as:
\begin{equation}
\begin{aligned}
\mathbf{X}_{down} &= \text{MLP}_{down}(\text{LN}(\mathbf{X})), \\
    \mathbf{X}_{mid} &= \text{MLP}_{mid}(\mathbf{X}_{down})+\mathbf{X}_{down} , \\
    \mathbf{X}_{up} &= \text{MLP}_{up}(\mathbf{X}_{mid})+\mathbf{X},
    \label{equ:umlp}
\end{aligned}            
\end{equation}
where $\text{MLP} (\cdot)$  consists of a linear layer and a $\text{GELU}$ activation \cite{hendrycks2016gaussian},  $\text{LN}$ denotes the Layer Normalization \cite{ba2016layer}. 

In the Spatial UMLP pathway, given that the number of joints ($K=17$) is already sufficiently sparse, we implement the operations in reverse order along the spatial dimension, starting with $\mathbf{X}_{up}$, followed by $\mathbf{X}_{mid}$, and finally $\mathbf{X}_{up}$.
This parallel design along both channel and spatial dimensions allows the UMLP module to extract rich feature representations across different scales and levels.

\begin{figure}[!t]
\centering
\centerline{\includegraphics[width=0.8\linewidth]{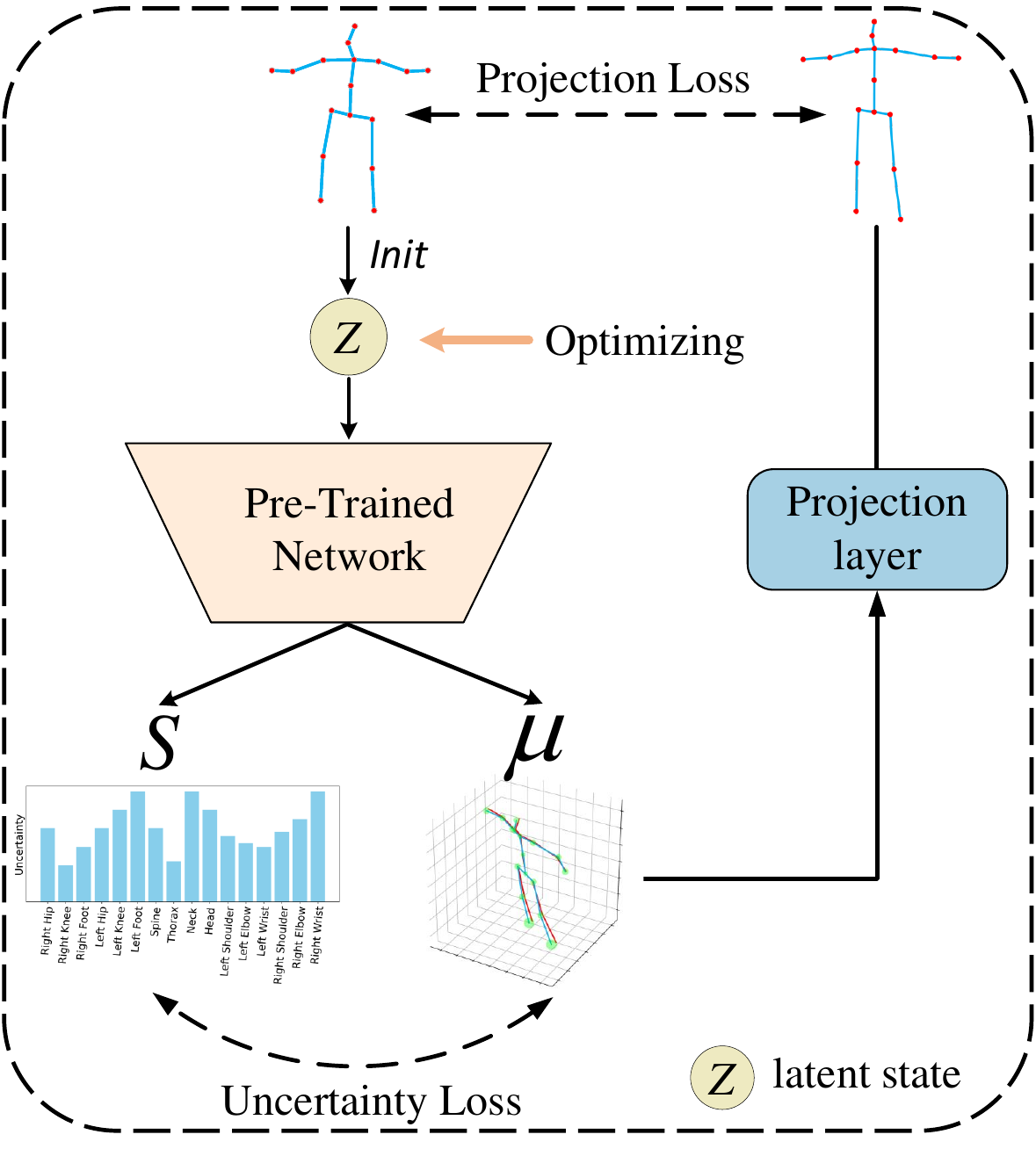}}
\caption{
Structure of the UAO framework. During the testing-time optimization, the parameters of the pre-trained model (GUMLP) are freezed. 
The variable being optimized is the latent state $\mathbf{z}$, which is initialized with the original 2D pose input $\mathbf{J}^{2D}$.
We use the pre-trained GUMLP to obtain the mean pose ${\pmb{\mu}}$ and uncertainty value $\pmb{s}$ of the corresponding 3D pose distribution. The predicted mean pose is projected back to 2D space and compared with the input pose $\mathbf{J}^{2D}$ to calculate the projection loss. The uncertainty loss is then calculated using ${\pmb{\mu}}$ and $\pmb{s}$. Under the constraints of projection and uncertainty, the latent state $\mathbf{z}$ is optimized during the iterations.
}
\label{fig:optimization}
\end{figure}

\section{Testing-Time Optimization}

\label{sec:optimization}

\subsection{Optimization Strategy}
\label{sec:optimization_strategy}

The proposed Uncertainty-Aware testing-time Optimization (UAO) framework is illustrated in Figure \ref{fig:optimization}.
It is essential to highlight that we only optimize the latent state $\mathbf{z} \in \mathbb{R}^{K\times 2}$ initialized by 2D input $\mathbf{J}^{2D}$. 
The parameters of the pre-trained model are fixed throughout the inference, which ensures the learned implicit prior from the training dataset not disrupted. Besides, we incorporate constraints into the optimization process. Specifically, projection constraint is applied to guarantee alignment of the optimized 3D pose in 2D space, and uncertainty constraint is introduced to determine the degree to which each joint can be optimized. After several iterations, the latent state $\mathbf{z}$ is refined progressively towards its optimal state. Ultimately, by feeding the optimized $\mathbf{z}^*$ into the pre-trained network, we obtain a high-quality 3D pose. The pseudo-code of the testing-time optimization is summarized in Algorithm \ref{algrothm1}.

\subsection{Projection Constraint}
Projection constraint is commonly used in optimization-based methods, such as SPIN \cite{kolotouros2019learning} and EFT \cite{joo2021exemplar}. For the $j$-th 3D pose $\mathbf{J}^{3D}_{j}$, its 2D projection $\mathbf{J}^{2D}_{j}$ should conform to the principles of perspective projection \cite{wang2014robust}, which can be expressed as follow:
\begin{equation}
\mathbf{J}^{3D}_{j} \cdot \mathbf{P} = \mathbf{J}^{2D}_{j},
\end{equation}
where the projection matrix $\mathbf{P} \in \mathbb{R}^{3 \times 2}$ contains the intrinsic parameters of the camera. In real scenarios, the intrinsic parameters of the camera can be obtained from camera specifications or can be inferred solely from the input images \cite{workman2015deepfocal}.
During the testing-time optimization, the 2D projection $\hat{\mathbf{J}}^{2D}_{j}=\hat{\mathbf{J}}^{3D}_{j} \cdot \mathbf{P}$ of the predicted 3D pose $\hat{\mathbf{J}}^{3D}_{j}$ is expected to closely align with the given 2D pose $\mathbf{J}^{2D}_{j}$. 
A large error between them indicates that the estimated 3D pose $\hat{\mathbf{J}}^{3D}_{j}$ may be problematic and needs to be corrected. 
The formulation of projection constraint $L^{opt}_{P,j}$ can be expressed as:
\begin{equation}
L^{opt}_{P,j} = ||\hat{\mathbf{J}}^{2D}_{j} - \mathbf{J}^{2D}_{j}||_{2}.
\end{equation}

The purpose of the projection constraint is to enforce the relationship between the estimated 3D pose $\hat{\mathbf{J}}^{3D}_{j}$ and its corresponding 2D projection $\hat{\mathbf{J}}^{2D}_{j}$ to follow the geometric principles of perspective projection, thereby improving the accuracy of the pose estimation.

\begin{algorithm}[t]
    \caption{Optimization Strategy}
    \label{algrothm1}
    \KwIn{
    2D pose $\mathbf{J}_j^{2D}$, pre-trained model $f^*$ with fixed parameters,
    projection matrix $\mathbf{P}$, optimization steps $T$
    }
    \KwOut{
    3D pose $\hat{\mathbf{J}}_j^{3D}$
    }    
    Initialization: $\mathbf{z}_j = \mathbf{J}_j^{2D}$, $iter=0$ \\
    $\hat{\pmb{\mu}}_j = f_{\pmb{\mu}}^*(\mathbf{J}_j^{2D}), \hat {\pmb{s}}_j = f_{\pmb{s}}^*(\mathbf{J}_j^{2D})$ \\
    \While{ $iter < T$}
    {
    $\hat{\mathbf{J}}_j^{3D} = f_{\pmb{\mu}}^*(\mathbf{z}_j)$ \\
    $L_{P,j}^{opt} = \|\hat{\mathbf{J}}_j^{3D} \cdot \mathbf{P}  - \mathbf{J}_j^{2D}\|_{2}$ \\
    $L_{U,j}^{opt} = \left\| \hat{\pmb{\mu}}_j -\hat{\mathbf{J}}_j^{3D} \right\|^2 / 2 \exp (\hat {\pmb{s}}_j)$\\
    $L_{total,j}^{opt} = \lambda_{P}\cdot L_{P,j}^{opt} + \lambda_{U}\cdot L_{U,j}^{opt}$ \\ 
    $\mathbf{z}_j \leftarrow Adam(\mathbf{z}_j, \nabla L_{total,j}^{opt})$;\\
    $iter = iter + 1$ \\
}
$\hat{\mathbf{J}}_j^{3D} = f_{\pmb{\mu}}^*(\mathbf{z}_j) $ \\
return $\hat{\mathbf{J}}_j^{3D}$
\end{algorithm}

\begin{table*}[t]
\caption{Quantitative comparison with the state-of-the-art methods on Human3.6M under Protocol 1. 
    Top-table: 2D pose detected by cascaded pyramid network (CPN) \cite{chen2018cascaded} is used as input. 
    Bottom-table: Ground truth 2D pose is used as input. 
    The top two best results for each action are highlighted in bold and underlined, respectively.
}
\resizebox{\textwidth}{!}{
\begin{tabular}{l|ccccccccccccccc|c}
\toprule
Method & Dire. & Disc. & Eat & Greet & Phone & Photo & Pose & Purch. & Sit & SitD. & Smoke & Wait & WalkD. & Walk & WalkT. & Avg   \\
\midrule
SimpleBaseline \cite{martinez2017simple} &51.8 & 56.2& 58.1& 59.0&69.5&78.4& 55.2& 58.1& 74.0 & 94.6 & 62.3 & 59.1 & 65.1& 49.5 & 52.4 & 62.9\\
VideoPose3D \cite{pavllo20193d}  & 47.1&50.6&49.0 &51.8 &53.6&61.4 &49.4 &47.4& 59.3 &67.4& 52.4& 49.5& 55.3 &39.5 &42.7 &51.8\\
LCN \cite{ci2019optimizing}  & 46.8 & 52.3 & \underline{44.7} & 50.4 & 52.9 & 68.9 & 49.6 & 46.4 & 60.2 & 78.9 & 51.2 & 50.0 & 54.8 & 40.4 & 43.3 & 52.7\\
SRNet \cite{zeng2020srnet}  & 44.5 & 48.2 & 47.1 & \textbf{47.8} & 51.2 & \underline{56.8} & 50.1 & 45.6 & 59.9 & 66.4 & 52.1 & \textbf{45.3} & 54.2 & 39.1 & \underline{40.3} & 49.9 \\
GraphSH \cite{xu2021graph}& 45.2 & 49.9 & 47.5 & 50.9 & 54.9 & 66.1 & 48.5 & 46.3 & 59.7 & 71.5 & 51.4 & 48.6 & 53.9 & 39.9 & 44.1 & 51.9 \\
MGCN \cite{zou2021modulated} & 45.4 & 49.2 &  45.7 & 49.4 & 50.4 & 58.2 & 47.9 & 46.0 & 57.5 & 63.0 & 49.7 & 46.6 & 52.2 & 38.9 & 40.8 & \underline{49.4} \\
GraFormer \cite{zhao2022graformer} & 45.2 &50.8 & 48.0 & 50.0 & 54.9& 65.0 & 48.2& 47.1&60.2&70.0&51.6&48.7&54.1&39.7&43.1& 51.8 \\
UGRN \cite{li2023pose} & 47.9 & 50.0 & 47.1 & 51.3 & 51.2 & 59.5 & 48.7 & 46.9 & \underline{56.0} & \underline{61.9} & 51.1 & 48.9 & 54.3 & 40.0 & 42.9 & 50.5 \\
MLP-JCG \cite{tang2023mlp} & \underline{43.8} & \textbf{46.7} & 46.9 & \underline{48.9} & \underline{50.3} & 60.1 & \textbf{45.7} & \underline{43.9} & \underline{56.0} & 73.7 & \underline{48.9} &48.2 & \underline{50.9} &39.9 &41.5 &49.7 \\
DiffPose \cite{gong2023diffpose}   & - & - & - & - & - & - & - & - & - & - & - & - & - & - & - & 49.7 \\
\midrule
GUMLP (Ours) & 44.3 & 49.8 & \underline{44.7} & 49.3 & 51.4 & 58.1 & 47.3 & 45.6 & 58.2 & 63.7 & 51.0 & 47.6 & 53.4 & \underline{38.1} & 40.4 & \underline{49.4} \\
GUMLP + UAO (Ours) & \textbf{42.3} & \underline{47.5} & \textbf{42.8} & \textbf{47.8} & \textbf{48.7} & \textbf{55.8} & \underline{45.8} & \textbf{42.4} & \textbf{55.2} & \textbf{60.9} & \textbf{48.6} & \underline{45.9} & \textbf{50.2} & \textbf{36.0} & \textbf{38.2} & \textbf{46.7}\\
\toprule
Method & Dire. & Disc. & Eat & Greet & Phone & Photo & Pose & Purch. & Sit & SitD. & Smoke & Wait & WalkD. & Walk & WalkT. & Avg   \\
\midrule
SimpleBaseline \cite{martinez2017simple}  &37.7 &44.4 &40.3 &42.1 &48.2 &54.9 &44.4 &42.1 &54.6 &58.0 &45.1 &46.4 &47.6 &36.4 & 40.4& 45.5\\
LCN \cite{ci2019optimizing} &36.3 &38.8 &\underline{29.7} &37.8 & 34.6 &42.5 &39.8 &32.5 &\underline{36.2} & \underline{39.5} &34.4 &38.4 &38.2 &31.3& 34.2 &36.3\\
MGCN \cite{zou2021modulated} & - & - & - & - & - & - & -  & - & - & - & - & - & - & - & - & 37.4 \\
GraFormer \cite{zhao2022graformer} & 32.0 & 38.0 & 30.4 & 34.4 & 34.7 &43.3 & 35.2 & 31.4 & 38.0 & 46.2 & 34.2 & 35.7 & 36.1 & \underline{27.4} & 30.6 & 35.2 \\
MLP-JCG \cite{tang2023mlp} & \textbf{29.1} & \underline{36.0} & 30.4 & 33.8 & 35.5 & 46.5 & 35.3 & \underline{31.2} & 39.2 & 48.8 &   \underline{33.9} & \underline{35.2} & \underline{35.8} & \textbf{26.9} & \underline{29.4} & 35.1 \\
DAF-DG \cite{peng2024dual}   & - & - & - & - & - & - & - & - & - & - & - & - & - & - & - & 44.4 \\
\midrule
GUMLP (Ours) & 30.2 & 36.2 & 30.3 & \underline{32.9} & \underline{33.6} & \underline{41.6} & \underline{35.1} & 31.8 & 38.7 & 42.6 & 34.8 & 36.8 & 37.2 & 28.7 & 30.0 & \underline{34.7}   \\
GUMLP + UAO (Ours) & \underline{29.5} & \textbf{34.5} & \textbf{27.5} & \textbf{31.1} & \textbf{31.8} & \textbf{38.4} & \textbf{34.0}  & \textbf{29.6} & \textbf{35.5} & \textbf{38.7} & \textbf{32.9} & \textbf{35.0} & \textbf{34.9} & 28.1 & \textbf{29.2} & \textbf{32.7}\\
\bottomrule
\end{tabular}
}
\label{tab:human3.6M_cpn}
\end{table*}

\subsection{Uncertainty Constraint}
Although the projection constraint can ensure the alignment of the generated pose in 2D space, it often brings over-fitting problems. This is due to the fact that multiple 3D poses might correspond to the same 2D pose after projection, which is an inherent challenge in the 3D human pose estimation task. 
When we solely depend on the projection constraint for the optimization process, 
we observe two phenomena: 1) some joints that were well-estimated deviate away from the ground truth; 2) some poorly estimated joints are initially close to the ground truth but then move away from it.
To alleviate this, we introduce uncertainty constraint, which allows the well-estimated joints to stay in their original positions and mainly optimize the joints with high uncertainties.

During the testing process, for the $j$-th 2D input, the pre-trained model provides the initial 3D result $\hat{\pmb{\mu}}_{j,k}$ and uncertainty  $\hat{\pmb{s}}_{j,k}$ associated with the  $k$-th estimated 3D joint. We can explicitly formulate the uncertainty constraint $L_{U,j}^{opt}$ as:
\begin{equation}
L_{U,j}^{opt}=\frac{1}{2K} \sum_{k=1}^K \left\|
\hat{\pmb{\mu}}_{j,k} - f_{\pmb{\mu}}^*(\mathbf{z})_{j,k} \right\|^2 \exp(-\hat{\pmb{s}}_{j,k}),
\end{equation}
where $f^*$ denotes the pre-traind model, $f_{\pmb{\mu}}^{*}$ denotes the predicted 3D result.
This constraint enables joints with higher uncertainty less constrained by the results of the initial estimation, allowing more room for optimization. On the other hand, the joint with lower uncertainty means that the network is confident about the result, so limits it from further drifting.

\subsection{Combined Optimization Loss Function}

The overall optimization loss $L_{total,j}^{opt}$ for the $j$-th case is defined as the weighted sum of the projection loss and the uncertainty loss:
\begin{equation}
    \L_{total,j}^{opt} = \lambda_{P}\, L_{P,j}^{opt} + \lambda_{U}\, L_{U,j}^{opt},
\end{equation}
where \(\lambda_{P}\) and \(\lambda_{U}\) are hyperparameters for each loss term.

\section{Experiments}

\subsection{Datasets and Evaluation Metrics}

Here, we provide a more detailed description of the datasets and evaluation metrics.

\noindent \textbf{Human3.6M} \cite{h36m} is the most representative benchmark for the estimation of 3D human poses.
It contains 3.6 million video frames captured from four synchronized cameras at 50Hz in an indoor environment.
There are 11 professional actors performing 17 actions, such as greeting, phoning, and sitting.
Following previous works \cite{pavllo20193d, cai2019exploiting}, we train our model on five subjects (S1, S5, S6, S7, S8) and test it on two subjects (S9 and S11). 
The performance is evaluated by two common metrics: MPJPE (Mean Per-Joint Position Error), termed Protocol 1, is the mean Euclidean distance between the predicted joints and the ground truth in millimeters. 
PA-MPJPE, termed Protocol 2, measures the MPJPE after Procrustes Analysis (PA) \cite{gower1975generalized}.

\noindent \textbf{MPI-INF-3DHP} \cite{mehta2017monocular} is a more challenging 3D pose dataset that contains both indoor and complex outdoor scenes.
There are 8 actors performing 8 actions from 14 camera views, which cover a greater diversity of poses. 
Its test set consists of three different scenes: studio with green screen (GS), studio without green screen (noGS), and outdoor scene (Outdoor). Following \cite{zou2021modulated,zeng2021learning,pavllo20193d}, we use Percentage of Correct Keypoints (PCK) with a threshold of 150mm and the Area Under Curve (AUC) for a range of PCK thresholds for evaluation.

\noindent \textbf{3DPW} \cite{pw3d} is a complex in-the-wild dataset that provides 3D pose and mesh annotations, obtained using hand-held cameras and IMUs. It comprises 60 videos recorded at 30 frames per second. To ensure a fair comparison, we train our model on Human3.6M and test it on the 3DPW training set.

\subsection{Implementation Details}

We implement our approach with PyTorch, training and testing it on one NVIDIA RTX 3090 GPU. 
Following \cite{zou2021modulated,cai2019exploiting}, we use 2D pose detected by CPN \cite{chen2018cascaded} for Human3.6M, and ground truth 2D pose for MPI-INF-3DHP.
During the training phase, we train a deep neural network for deep prior. The proposed GUMLP is designed by stacking GCN layers and UMLP modules for $N=3$ times.
For the testing time optimization, the hyperparameters for projection loss and uncertainty loss are chosen as $\lambda_{P}=1$ and $\lambda_{U}=0.005$, respectively. 

\begin{table*}[!h]
\caption{
Quantitative comparison with the state-of-the-art methods on Human3.6M under Protocol 2. 
SMPLify-X \cite{pavlakos2019expressive} uses 2D pose detected by OpenPose \cite{cao2017realtime} as input, whereas other methods use 2D poses detected by cascaded pyramid network (CPN) \cite{chen2018cascaded}.
$\S$ denotes the methods that use refinement module \cite{zou2021modulated, cai2019exploiting}.
The top two best results for each action are highlighted in bold and underlined, respectively.}
\resizebox{\textwidth}{!}{
\begin{tabular}{l|ccccccccccccccc|c}
\toprule
Protocol 2 & Dire. & Disc. & Eat & Greet & Phone & Photo & Pose & Purch. & Sit & SitD. & Smoke & Wait & WalkD. & Walk & WalkT. & Avg   \\
\midrule
SMPLify-X \cite{pavlakos2019expressive} & - & - & - & - & - & - & - & - & - & - & - & - & - & - & - & 75.9 \\
SimpleBaseline \cite{martinez2017simple} & 39.5 & 43.2 & 46.4 & 47.0 & 51.0 & 56.0 & 41.4 & 40.6 & 56.5 & 69.4 & 49.2 & 45.0 & 49.5 & 38.0 & 43.1 & 47.7\\
VideoPose3D \cite{pavllo20193d}  & 36.0 & 38.7 &38.0 &41.7 &40.1 &45.9 &37.1 &35.4 &46.8 &53.4 &41.4 &36.9 &43.1 &30.3 & 34.8 & 40.0 \\
LCN \cite{ci2019optimizing}  &36.9 &41.6 &38.0 &41.0& 41.9 &51.1 &38.2 &37.6 &49.1 &62.1 &43.1 &39.9 &43.5 &32.2 &37.0 & 42.2 \\
STGCN \cite{cai2019exploiting}$\S$ & 36.8 & 38.7 & 38.2 & 41.7 & 40.7 & 46.8 & 37.9 & 35.6 & 47.6 & \textbf{51.7} & 41.3 & 36.8 & 42.7 & 31.0 & 34.7 & 40.2 \\
SRNet \cite{zeng2020srnet} &35.8 &39.2& 36.6 &\textbf{36.9} &39.8 &45.1 &38.4& 36.9 & 47.7 & 54.4 & \textbf{38.6} & 36.3& \textbf{39.4} &30.3 &35.4& 39.4\\
MGCN \cite{zou2021modulated}$\S$ & 35.7 & 38.6 & 36.3 & 40.5 & \textbf{39.2} & \underline{44.5} & 37.0 & 35.4 & 46.4 & 51.2 & 40.5 & \textbf{35.6} & 41.7 & 30.7 & 33.9 & \underline{39.1} \\
MLP-JCG \cite{tang2023mlp} & \textbf{33.7} & \textbf{37.4} & 37.3 & \underline{39.6} & 39.8 &47.1 &\textbf{33.7}&\textbf{33.8} &\textbf{45.7} &60.5& \underline{39.7} & 37.7& \underline{40.1} & \underline{30.1} & \underline{33.8} & 39.3 \\
GKONet \cite{hu2023geometric} & 35.4 &38.8 & \underline{35.9} &40.4 &\underline{39.6} & \textbf{44.0} &36.7 &35.4& 46.8 &53.7 & 40.9 & 36.6 & 42.0 & 30.6 & 33.9 & 39.4 \\
ZEDO \cite{jiang2024back}  & - & - & - & - & - & - & - & - & - & - & - & - & - & - & - & 42.1\\
\midrule
GUMLP (Ours) & 35.3 & 38.5 & \underline{35.9} & 40.2 & 40.3 & 44.9 & \underline{36.3}  & 35.0 & \underline{46.2} & \underline{51.8} & 41.2 & \underline{35.8} & 41.7 & 30.3 & 33.9 & 39.2 \\
GUMLP + UAO (Ours) & \underline{34.9} & \underline{38.3} & \textbf{35.3} & 40.2 & 40.1 & 44.9 & 36.5 & \underline{34.8} & \underline{46.0} & 51.9 & 40.8 & 36.0 & 42.3 & \textbf{30.0} & \textbf{33.4} & \textbf{39.0} \\
\bottomrule
\end{tabular}
}
\label{tab:human3.6M_cpn_P2}
\end{table*}

\subsection{Comparison with State-of-the-art Methods}
\noindent \textbf{Human3.6M.}
The comparison results with other state-of-the-art methods under Protocol 1 are shown in Table \ref{tab:human3.6M_cpn}. 
Following \cite{zou2021modulated,cai2019exploiting, xu2021graph}, the 2D pose detected by CPN \cite{chen2018cascaded} is used as input for training and testing.
We have observed that our baseline model, GUMLP, produces competitive results.
It is worth noting that none of the compared methods employed a post-optimization approach.
To further enhance the accuracy of our model predictions, we integrate the Uncertainty-Aware Optimization (UAO) framework, which leverages both projection and uncertainty constraints. This integration significantly improved GUMLP's performance, reducing the error from 49.4mm to 46.7mm in MPJPE, \textbf{a relative improvement of 5.5\%}.
For Protocol 2, we also obtain the best overall results, as shown in Table \ref{tab:human3.6M_cpn_P2}.
These results validate the effectiveness of the designed 2D-to-3D lifting network (GUMLP) and the UAO framework.
To further explore the lower bounds of our approach, we compare it with previous state-of-the-art methods with ground truth 2D pose as input.
As shown in Table \ref{tab:human3.6M_cpn} (bottom), we achieve state-of-the-art performance, outperforming all other methods.
Additionally, we observe that the performance of GUMLP can be further enhanced with the UAO framework, even when the latent state $\mathbf{z}$ is initialized with the ground truth 2D pose.
This improvement is attributed to the limited capacity of the pre-trained GUMLP and the effective enhancement of 3D pose quality through the application of projection and uncertainty constraints during the optimization process. 
This indicates hat optimizing the latent state $\mathbf{z}$ aims to refine the 3D pose, rather than simply making $\mathbf{z}$ closer to the ground truth 2D pose.

\begin{table}[!t]
    \centering
     \caption{Quantitative comparisons with state-of-the-art methods on the MPI-INF-3DHP dataset.}
    \setlength\tabcolsep{1.30mm}
    \footnotesize
    \resizebox{1.0 \linewidth}{!}{
    \begin{tabular}{l|c|c|c|c|c}
    \toprule 
    Method  & GS $\uparrow$ & noGS $\uparrow$&  Outdoor $\uparrow$ & All PCK $\uparrow$& All AUC $\uparrow$\\
    \midrule
    SimpleBaseline \cite{martinez2017simple} & 49.8 & 42.5 & 31.2 &42.5 & 17.0 \\
    GraphSH \cite{xu2021graph} & 81.5 & 81.7 & 75.2 & 80.1 & 45.8 \\ 
        MGCN \cite{zou2021modulated} & 86.4 & 86.0 & 85.7 & 86.1 & 53.7  \\
    GraFormer \cite{zhao2022graformer} & 80.1 &77.9 &74.1 &79.0& 43.8 \\
    UGRN \cite{li2023pose} & 86.2 & 84.7 & 81.9 & 84.1 & 53.7 \\
    \midrule
    GUMLP & \underline{87.6} & \underline{85.9} & \underline{89.3} & \underline{87.4} & \underline{54.8}  \\
    GUMLP + UAO & \textbf{88.0} & \textbf{86.9} &\textbf{89.8} & \textbf{88.0} & \textbf{55.1}  \\
    \bottomrule
    \end{tabular}
    }
    \label{tab:3dhp}
\end{table}

\noindent \textbf{MPI-INF-3DHP.}
To evaluate the generalization ability of the proposed UAO framework, we further compare our approach against previous state-of-the-art methods on cross-dataset scenarios. 
As shown in Table \ref{tab:3dhp}, GUMLP obtains the best results in all scenes and all metrics, consistently surpassing other methods.
After utilizing the UAO framework, the performance can be further improved in both indoor and outdoor scenarios. This suggests that the designed testing-time optimization process can effectively enhance the model's ability to generalize to unknown actions and datasets.

\begin{table}[!t]
    \footnotesize
    \caption{Cross-domain evaluation on 3DPW. Source dataset: Human3.6M, target dataset: 3DPW.}
    \centering
    \setlength{\tabcolsep}{3.2mm}
\begin{tabular}{l|cc}
\toprule
    Method & PA-MPJPE $\downarrow$ &  MPJPE $\downarrow$ \\
\midrule
    ISO \cite{zhang2020inference} & 70.8 & - \\
    Wang \textit{et al.} \cite{wang2020predicting} & 68.3 & 109.5 \\
    PoseAug \cite{gong2021poseaug}         &    58.5    &  94.1  \\
    PoseDA \cite{chai2023global}  &  55.3   &  87.7 \\
    PMCE  \cite{you2023co}              &  52.3   &  81.6 \\
    AdaptPose \cite{gholami2022adaptpose}  &    46.5        & 81.2   \\
     ZEDO \cite{jiang2024back}           &  42.6   &  80.9 \\ 
    BOA \cite{guan2021bilevel}          &  49.5   & 77.2   \\
    DynaBOA \cite{guan2022out}    & \underline{40.4}  & 65.5 \\
    DAPA \cite{weng2022domain}       & 46.5    & 75.0  \\
    CycleAdapt \cite{nam2023cyclic}   & \textbf{39.9}  & \underline{64.7} \\ 
    ESCAPE \cite{bidulka2025escape}         & 47.9  &  79.0 \\
\midrule
GUMLP (Ours)   &  42.0 & 67.5 \\
GUMLP + UAO (Ours)  & 41.1  &  \textbf{60.8}  \\
\bottomrule
\end{tabular}
\label{performance_3dpw}
\end{table}

\noindent \textbf{3DPW.}
We further evaluated our model's performance on the 3DPW dataset, which features complex outdoor scenes. As shown in Table \ref{performance_3dpw}, our baseline model, GUMLP, outperforms previous methods in both PA-MPJPE and MPJPE metrics. Furthermore, the integration of the UAO framework improves performance by reducing MPJPE by 6.7mm. This result highlights the effectiveness of our proposed optimization strategy in outdoor environments.

\subsection{Ablation Study}

\noindent \textbf{Impact of Model Design.}
The GUMLP model primarily comprises two components: a GCN layer and a parallel U-shaped multi-layer perception (UMLP) module, as illustrated in Figure \ref{fig:model}. 
To investigate the design choices of GUMLP, experiments were conducted on Human3.6M using the 2D poses extracted by CPN \cite{chen2018cascaded} as input. The results are reported in Table \ref{tab:ab_study_GUMLP}. Our findings indicate that using only the spatial UMLP, which captures multi-scale features, or only the channel UMLP, which captures multi-level features, does not yield the best performance. The optimal results are achieved when both modules are used together, demonstrating that a combination of multi-level and multi-scale features is necessary for optimal performance.

\noindent \textbf{Impact of Model Configurations.}
We conducted ablation studies by exploring the performance of GUMLP with different hyperparameters.
The results are presented in Table \ref{table:diff_layers}.
We observe that with a fixed number of layers $L$, the model consistently achieves optimal performance with a hidden size $C$ of 512. Doubling or halving the hidden size results in performance degradation.This is because too few parameters lead to underfitting, while too many cause overfitting and hinder convergence.
With a hidden size $C$ of 512, we determined that the optimal number of layers is $L=3$  for the best performance.
In Table \ref{table:diff_scale}, we further explore the performance of Spatial UMLP with different spatial dimension $S_{mid}$ after the up-projection layer. The results show that when $S_{mid}=17*6$, our model achieves the best performance.

\begin{table}  
    \footnotesize
    \centering
    \caption{
    Ablation study on different designs of our model.
    }
    \setlength\tabcolsep{2.6mm}
    \footnotesize
    \begin{tabular}{c|cc|c}    
    \toprule
     GCN        &  Channel UMLP & Spatial UMLP & MPJPE $\downarrow$ \\
    \midrule
     \cmark     &      \cmark   &              & 50.5  \\ 
      \cmark    &               &    \cmark    & 56.2  \\ %
    \midrule
    \cmark      &     \cmark    &    \cmark    &  49.4  \\ %
    \bottomrule
    \end{tabular}    
    \label{tab:ab_study_GUMLP}
\end{table}

\begin{table}[htbp]
\footnotesize
\caption{Ablation study on various configurations of GUMLP. $L$ denotes the number of layers, $C$ denotes the hidden size.  }
    \centering
    \setlength{\tabcolsep}{4.1mm}
\begin{tabular}{l|cccc}
\toprule
L    &   C   & Params (M) &  FLOPs (M) & MPJPE $\downarrow$  \\  
\midrule
2    &  256  & 1.55   & 25.27  &  52.1 \\
2    &  512  &  5.91       & 76.44 &  50.8  \\
2    &  1024  &  23.33     &  256.49 &  54.0 \\
\hline
3    &  256  &  1.92     &   37.25  &  50.8 \\
3    &  512  &  7.32       &  112.1  &  \textbf{49.4}  \\
3    &  1024  &  28.87        &   374.61  &   50.6  \\
\hline
4    &  256  & 2.29      &   49.23  &  50.5 \\
4    &  512  & 8.72      &  147.76 &  \underline{49.6} \\
4    &  1024  & 34.41      &   492.72  &   51.4 \\
\bottomrule
\end{tabular}
\label{table:diff_layers}
\end{table}

\begin{table}[htbp]
\footnotesize
\caption{Ablation study on Spatial UMLP. $L$ denotes the number of layers, $C$ denotes the hidden size.  $S_{mid}$ denotes the spatial dimensions after the up-projection layer.}
    \centering
    \setlength{\tabcolsep}{3.1mm}
\begin{tabular}{l|ccccc}
\toprule
L    &   C   &  $S_{mid}$  & Params (M)      &  FLOPs (M)   & MPJPE $\downarrow$ \\
\hline
3    &  512  &     17 $\times$ 1   &   7.21           &   78.8      &  50.0 \\
3    &  512  &       17 $\times$ 3 &   7.23           &   86.7      &  50.1  \\
3    &  512  &       17 $\times$ 6&    7.32          &   112.1     &  \textbf{49.4}  \\
3    &  512  &       17 $\times$ 9&    7.45          &   153.1     &   50.0  \\
\bottomrule
\end{tabular}
\label{table:diff_scale}
\end{table}

\noindent \textbf{Impact of the Initialization Prior.}
In Table~\ref{tab:ab_initialization_prior_constraints_R1}, we use a template 2D pose to initialize the latent state \(\mathbf{z}\), treating this as a condition lacking an effective initialization prior. Rows 1 and 4 correspond to general data-driven approaches, where the 2D pose is directly fed into a fixed network to generate an estimated 3D pose without any further optimization.
Comparing rows 1–3 with rows 4–6, it is evident that a good initialization is crucial for achieving favorable convergence. Even the worst-case scenario of using the input 2D pose for initialization performs better than cases without an initialization prior. These results emphasize the significance of an effective initialization strategy in improving overall performance.

\begin{figure}[t]
    \centering
    \includegraphics[width=1.0\linewidth]{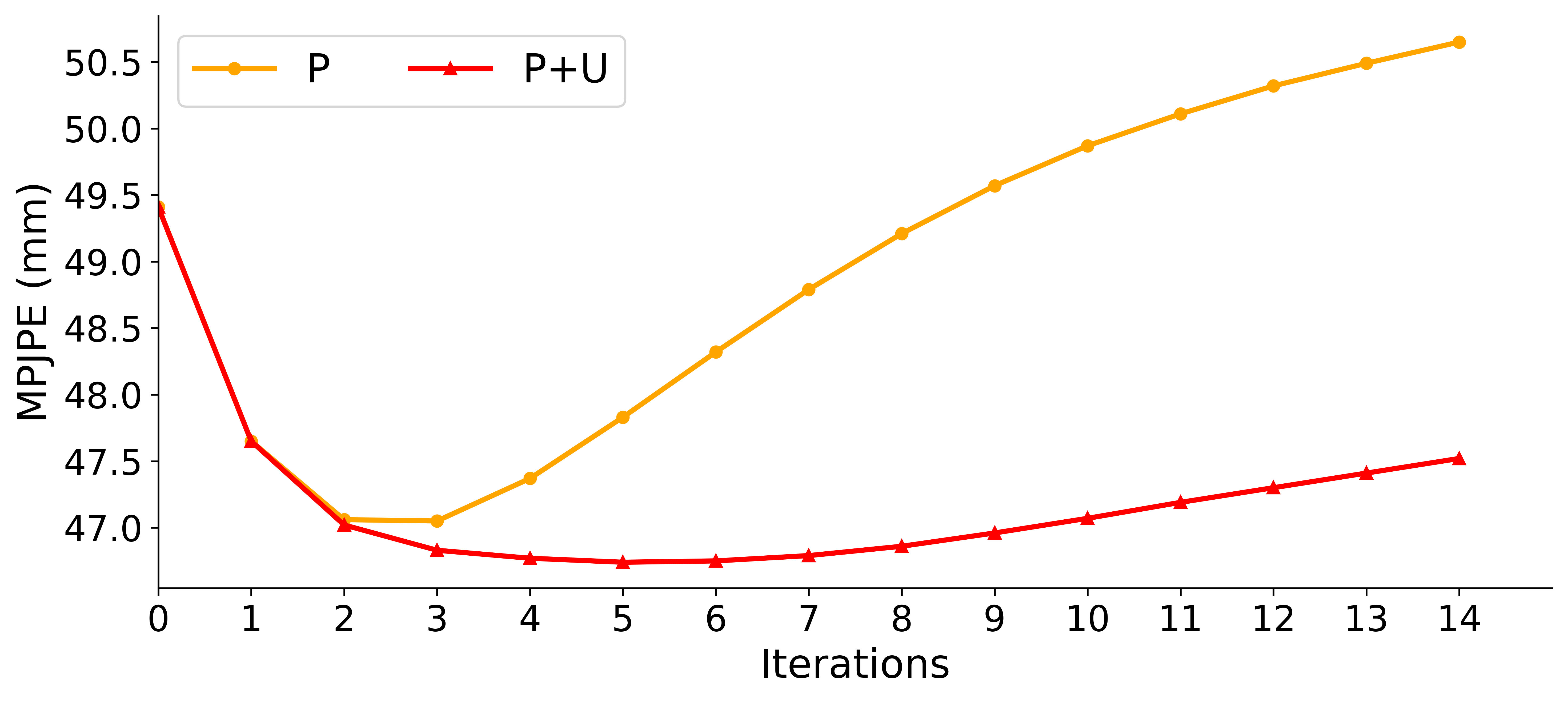}
    \caption{
    Performance with different iterations on Human3.6M test set. Here we adopt the pre-trained GUMLP as baseline and use the 2D pose detected by CPN \cite{chen2018cascaded} to initialize the latent state $\mathbf{z}$. $\mathbf{P}$ and $\mathbf{U}$ represent projection constraint and uncertainty constraint, respectively. 
    }
    \label{fig:plt_P_U}
\end{figure}

\noindent \textbf{Impact of Constraints during Optimization.}
Setting appropriate targets for test-time optimization is essential for achieving high-quality 3D poses. To this end, we evaluate the effectiveness of two proposed constraints in the optimization process.
Table \ref{tab:ab_initialization_prior_constraints_R1} presents the ablation results using the UAO framework on the Human3.6M test set with different constraints.
Comparing rows 2–3 with rows 5–6, the results show that the projection constraint significantly enhances the performance of the purely data-driven method. Furthermore, the introduction of the uncertainty constraint further improves the model’s generalization ability, leading to superior performance. These results highlight the effectiveness of both constraints in the optimization process.
To further explore the optimization mechanism, we illustrate the performance with different iterations on the test set of Human3.6M, as shown in Figure \ref{fig:plt_P_U}.
We observe that using only the projection constraint leads to a rapid decrease in MPJPE, followed by a gradual increase.
In contrast, the error can be stabilized at a lower level with the further introduction of the uncertainty constraint, even with more iterations.
The main reason is that the projection constraint only enforces consistency in the 2D plane, which is easy to overfit, yielding unrealistic poses in the 3D space.
To complement this, the uncertainty constraint helps maintain the output within a plausible 3D manifold. It restricts joints with low uncertainty from deviating too far from the ground truth while allowing joints with high uncertainty to have a larger divergence radius.
Experimental results demonstrate the robust adaptability of our method, especially in real-world scenarios where the optimal number of iterations is uncertain.

\begin{table}[t]  
    \footnotesize
    \centering
    \caption{Ablation study on initialization prior and constraints.}
    \setlength\tabcolsep{0.4mm}
    \footnotesize
    \begin{tabular}{c|c|ccc|c}    
    \toprule
Row  &   Initialization &  Initialization & Projection  & Uncertainty  & \multirow{2}{*}{MPJPE $\downarrow$ } \\
Index &   Manner           &    Prior                               &  Constraint  &  Constraint  &                       \\
    \midrule
     1   &\multirow{3}{*}{2D Template Pose} &    &             &             &     351.8    \\ 
     2   &                                        &    &   \cmark    &             &     93.1    \\ 
     3   &                                        &    &   \cmark    &   \cmark    &     69.5   \\ 
 \midrule
    4   & \multirow{3}{*}{Input 2D Pose}  & \cmark  &           &          &  49.4     \\  
    5   &                                       & \cmark  &   \cmark  &          &  47.1     \\  
    6    &                                & \cmark  &   \cmark  &  \cmark  &  46.7      \\  
    \bottomrule
    \end{tabular}
    \label{tab:ab_initialization_prior_constraints_R1}
\end{table}

\begin{figure}[b]
    \centering
    \includegraphics[width=1.0\linewidth]{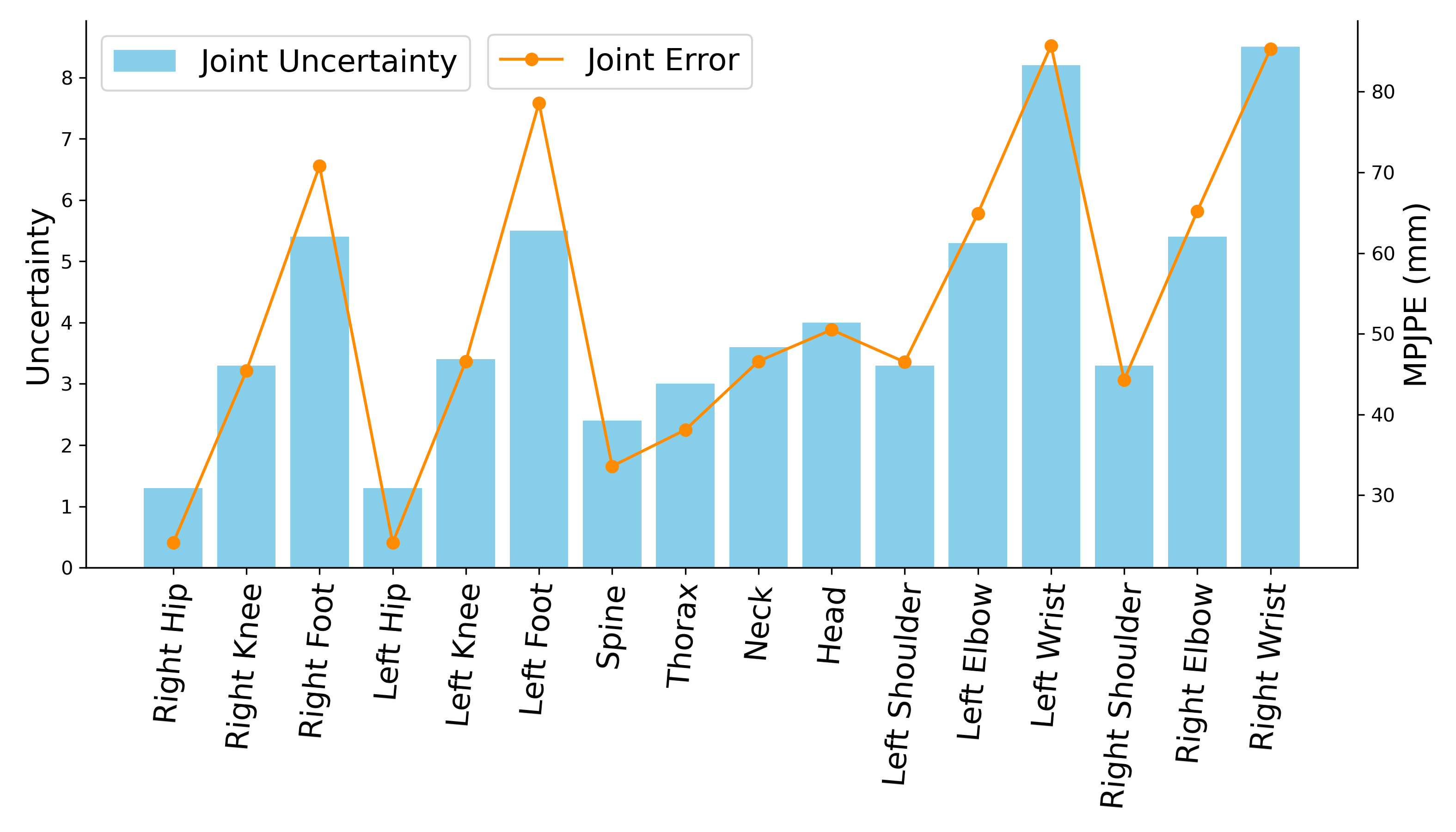}
    \caption{
    Combination of a histogram representing the uncertainty for each joint and a line chart depicting the mean error (MPJPE) for each joint.
    The left y-axis denotes the magnitude of uncertainty, while the right y-axis signifies the error value in MPJPE. }
    \label{fig:plt_Uncertainty_MPJPE}
\end{figure}

\begin{figure}[b]
\centering
\centerline{\includegraphics[width=0.9\linewidth]{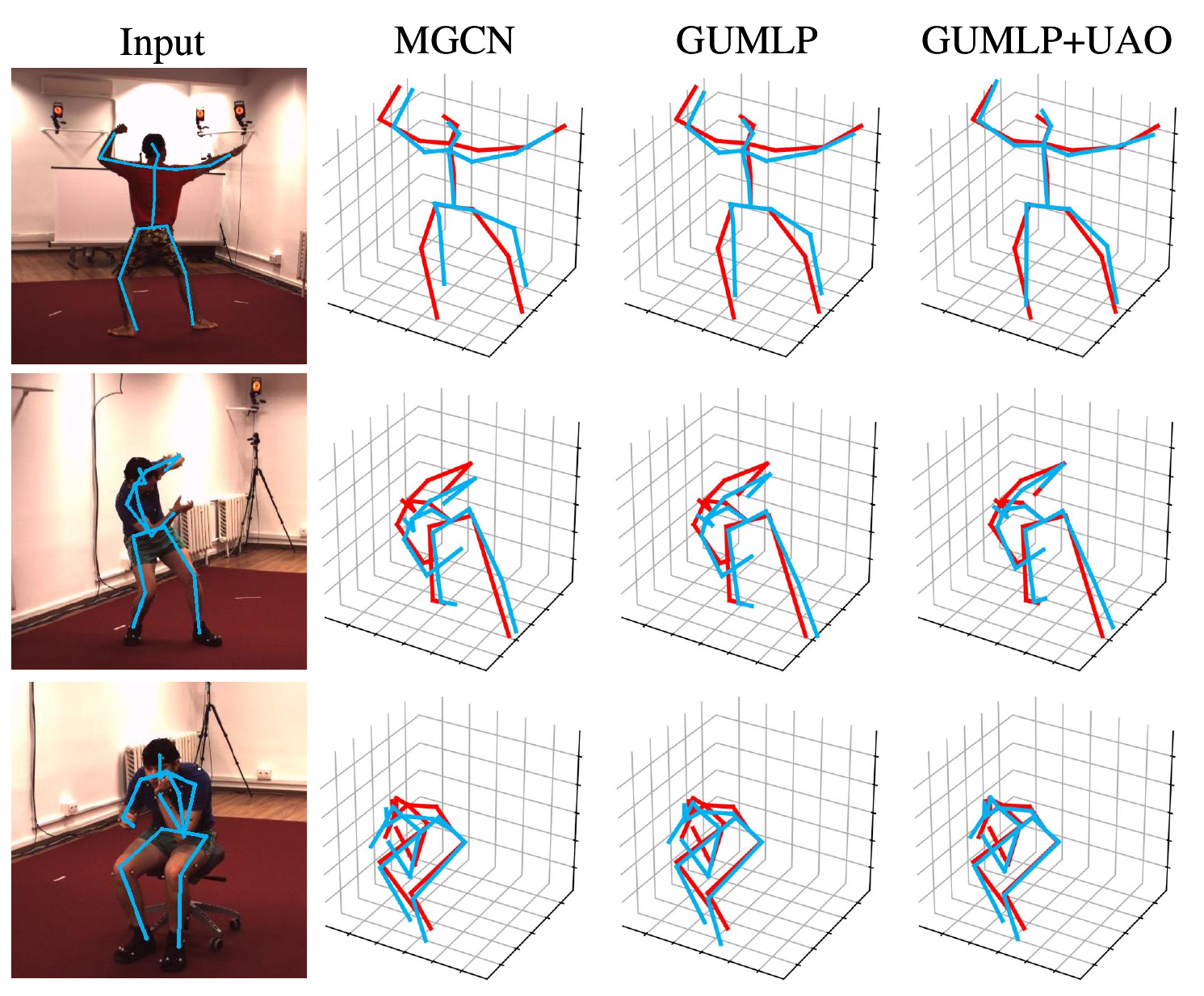}}
\caption
{Quantitative comparison of MGCN \cite{zou2021modulated}, GUMLP, and GUMLP with UAO on Human3.6M.
The blue pose represents the predicted results, while the red pose represents the ground truth.
}
\label{fig:qualitative_vis}
\end{figure}

\begin{figure*}[t]
\centering
  \includegraphics[width=1.0 \linewidth]{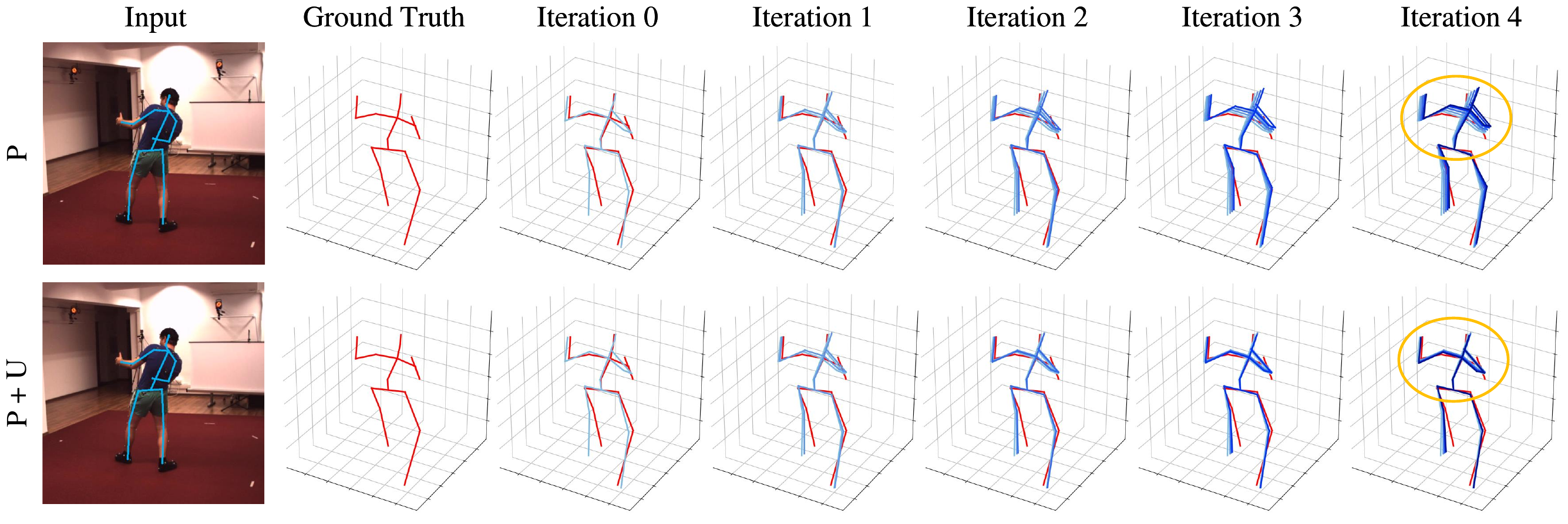}
    \caption{
    Visualization of the iterative optimization process. The red pose is the ground truth and the blue pose is our prediction. Color intensity reflects the number of iterations, with darker colors indicating more iterations. The initial 3D pose generated by the pre-trained model (column 3) progressively improves through iterative optimization (columns 4-7). ``P" indicates the use of only the projection constraint, ``P+U" denotes the combined use of projection and uncertainty constraints.
    }
\label{fig:pose_opt4}
\end{figure*}

\noindent \textbf{Inference speed of different Optimization Strategies.}
As is well known, optimization-based methods are more time-consuming.
We evaluated the inference speed of various optimization strategies on a single frame using one GeForce RTX 3090, with results shown in Table \ref{table:inference_speed_spin}. 
The baseline models include our GUMLP and the human mesh reconstruction method SPIN \cite{kolotouros2019learning}. 
Although these baselines differ, their inference times are relatively fast and have minimal impact on overall runtime. Since the primary computational cost lies in the optimization stage, we can fairly compare the runtime of different methods.
Our baseline model, GUMLP, reaches \textbf{310} fps. By setting the number of iterations to 5, our UAO framework reaches \textbf{48} fps,  surpassing all the other optimization methods.
This improvement is attributed to our UAO framework optimizing the latent state $\mathbf{z}$ rather than the entire model parameters, thereby accelerating the inference process.
It is worth noting that the output of any layer within the network can be selected as the latent state for optimization. For faster processing while maintaining satisfactory performance, we select the latent state before the final fully connected layer as the optimized variable, denoted as UAO*. This choice significantly enhances the inference speed, achieving up to \textbf{125} fps.

\begin{table}[t]
\footnotesize
\caption{ Inference speed of optimization methods. UAO* signifies employing the network's latent state as the variable for optimization. 
} 
    \centering
    \setlength{\tabcolsep}{9.1mm}
\begin{tabular}{l|c}
\toprule
Method & FPS \\
\midrule
GUMLP (Ours) & 310.0   \\
SPIN \cite{kolotouros2019learning} & 60.2 \\
\midrule
SPIN + BOA \cite{guan2021bilevel}       &  1.2 \\
SPIN + DynBOA \cite{guan2022out}        &  0.9    \\ 
SPIN + DAPA \cite{weng2022domain} & 2.3 \\
SPIN + CycleAdapt \cite{nam2023cyclic} & 13.5 \\
SPIN + ESCAPE \cite{bidulka2025escape} & 43.5 \\
\midrule
GUMLP + UAO (Ours)   &  \underline{48.0}  \\ 
GUMLP + UAO* (Ours)   & \textbf{125.0}   \\ 
\bottomrule
\end{tabular}
\label{table:inference_speed_spin}
\end{table}

\subsection{Uncertainty and Error for each Joint}

Figure \ref{fig:plt_Uncertainty_MPJPE} shows the average uncertainty and average error for each joint estimation across the entire test set, using the GUMLP model pre-trained on the human3.6M dataset.
For uncertainty estimation, we observe that the joints at the end of limbs often exhibit significant predicted uncertainty, such as the left wrist and the left foot.
This phenomenon is attributed to the fact that joints at the ends of the limbs typically experience higher movement velocities, presenting challenges in accurate prediction.
Moreover, the line chart illustrating joint errors follows a trend similar to the histogram, indicating that joints with larger uncertainties generally exhibit higher prediction errors. 
This further validates the effectiveness of GUMLP in predicting joint uncertainties.

\subsection{Generalization to Unseen Camera Views}

\begin{table}  
    \footnotesize
    \centering
    \caption{Performance across different views on Human3.6M in MPJPE. In-domain refers to testing under the same view as the training, while out-of-domain refers to testing under views other than the training view. $P_i$ denotes the $i$th camera view.   }
    \setlength\tabcolsep{1.3mm}
    \footnotesize
    \begin{tabular}{l|c|c|c|c|c|c}
    \toprule
     \multirow{2}{*}{Training views} & \multicolumn{3}{c|}{In Domain} & \multicolumn{3}{c}{Out-of-Domain}  \\
     \cline{2-7}
                                                    & GUMLP  & + UAO & $\Delta$ & GUMLP  & + UAO & $\Delta$ \\
    \midrule
        $P_1$ &  55.3 & 53.2 & 2.1 & 87.0  & 82.5 &  4.5 \\ 
        $P_1,P_2$ & 49.9  & 47.4  & 2.5 & 62.5 & 57.3 & 5.2 \\ 
        $P_1,P_2,P_3$ & 50.3 & 47.6 & 2.7 & 59.6 & 54.0 & 5.6 \\ 
        $P_1,P_2,P_3,P_4$ &  49.4 & 46.7 & 2.7 & -  & -& - \\ 
    \bottomrule
    \end{tabular}
    \label{tab:ab_study_views}
\end{table}

Table~\ref{tab:ab_study_views} presents the relative gains of the proposed approach over in- and out-domain cases on Human3.6M, where four camera views are available.
It can be inferred that the UAO framework can not only help the in-domain cases but also bring more significant improvements to unseen camera views. This verifies its ability to break the domain gap and meet our design expectations. Besides, from the last column of the table, a large training set can bring more optimization gains, which may be explained by better pre-training leading to a better depiction of the 3D pose manifold.

\subsection{Qualitative Results}

The visualization results on the Human3.6M dataset are shown in Figure \ref{fig:qualitative_vis}.
We can see that our original GUMLP is marginally superior to the previous state-of-the-art method, MGCN \cite{zou2021modulated}.
Moreover, employing the UAO framework brings better generation with more precise poses.
For example, the 3D pose of photoing action (row 2) predicted by GUMLP with UAO (col 4) is similar to the ground truth. As a comparison, the left arms predicted by MGCN and the GUMLP are at a lower position. 
We further visualize the iterative process of the UAO framework under different constraints, as shown in Figure \ref{fig:pose_opt4}. 
The results in Figure \ref{fig:pose_opt4} (row 1) show that relying solely on the projection constraint can cause some well-estimated joints to deviate from the ground truth.
However, for the same 2D pose, incorporating the uncertainty constraint (Figure~\ref{fig:pose_opt4} (row 2)) results in an optimized pose that better aligns with the ground truth. This further validates the effectiveness of the optimization process and highlights the necessity of our uncertainty constraint.

\begin{figure}[t]
  \centering
  \includegraphics[width=1.0 \linewidth]{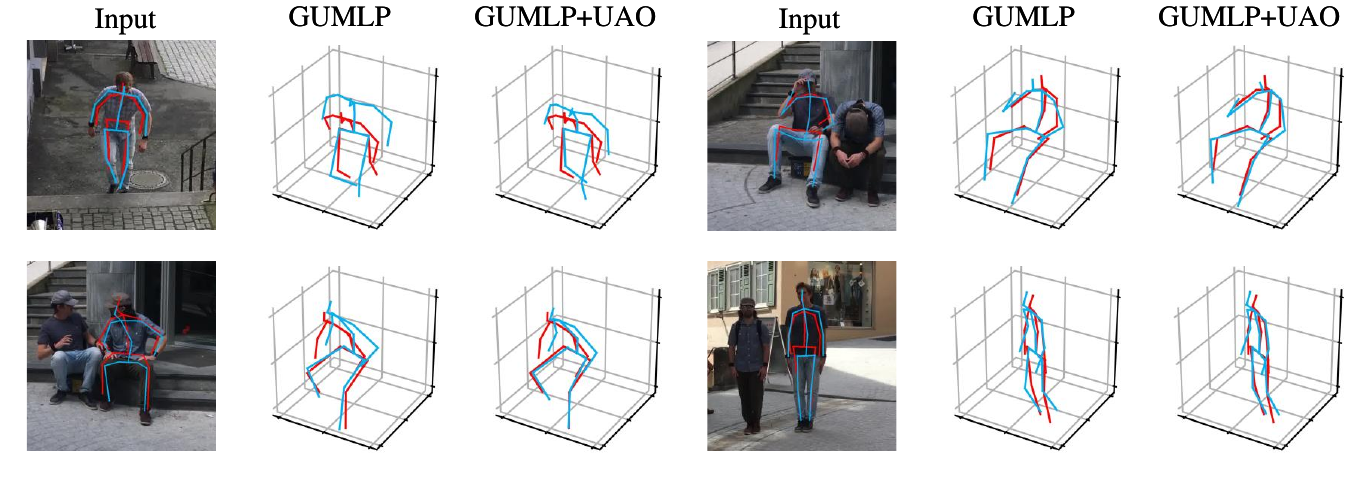}
    \caption{Visualization results on 3DPW using 2D pose detected by HRNet \cite{wang2020deep}. The red and blue poses in the Input image represent the ground truth and detected 2D poses, respectively. In the 3D pose columns (2–3), the red pose indicates the ground truth 3D pose, while the blue 3D pose corresponds to the predicted results based on the detected 2D input.
    }
  \label{pw3d_detected2D}
\end{figure}

\begin{figure}[t]
  \centering
  \includegraphics[width=1.0 \linewidth]{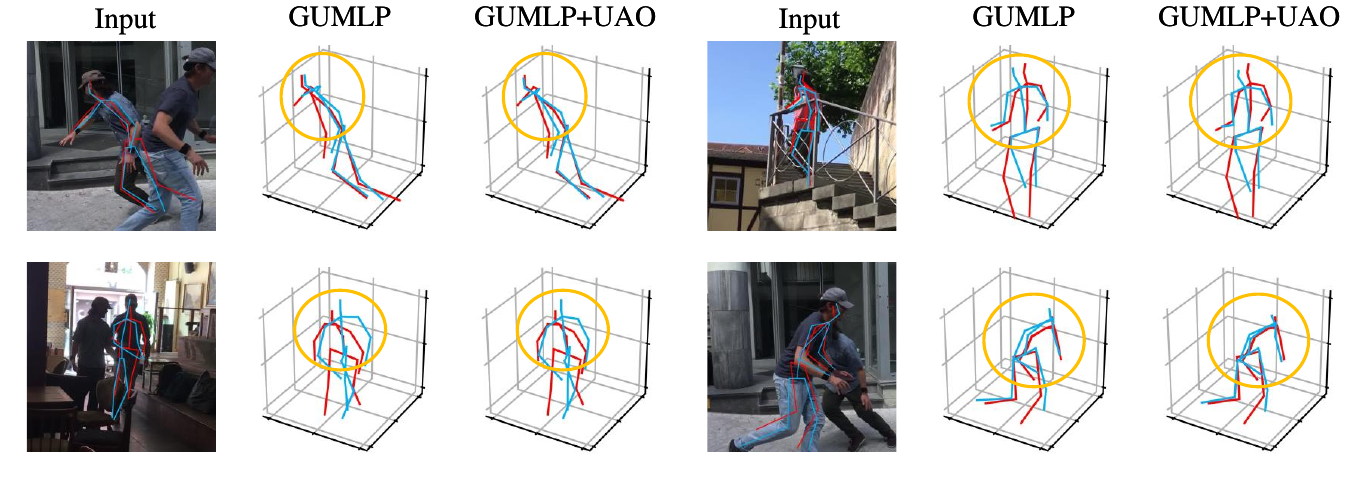}
    \caption{Occluded samples in 3DPW with 2D poses detected by HRNet \cite{wang2020deep}. The red and blue poses in the Input image represent the ground truth and detected 2D poses, respectively. In the 3D pose columns (2–3), the red pose indicates the ground truth 3D pose, while the blue 3D pose corresponds to the predicted results based on the detected 2D input.}
  \label{occlude_pose_detected}
\end{figure}

To further demonstrate the performance in outdoor scenes, we present the qualitative results on the 3DPW test set using 2D pose detected by HRNet \cite{wang2020deep}, as illustrated in Figure \ref{pw3d_detected2D}. The results indicate that the initial 3D outputs generated by GUMLP are significantly enhanced when using the UAO framework.
Figure \ref{occlude_pose_detected} also shows some occluded samples. Although our optimization strategy (UAO) is not specifically designed for occluded poses, it still performs well even in the presence of occluded body parts.

\section{Conclusion}
In this paper, we introduced an Uncertainty-Aware testing-time Optimization (UAO) framework for 3D human pose estimation. During training, our proposed GUMLP network estimates 3D poses along with uncertainty values for each joint. At test time, the UAO framework freezes the pre-trained network parameters and optimizes a latent state initialized by the input 2D pose. To effectively constrain the optimization in both 2D and 3D spaces, we apply projection and uncertainty constraints. 
Extensive experiments show that our approach achieves state-of-the-art results on popular datasets and demonstrates superior generalization in challenging outdoor scenes.

{\small
\bibliographystyle{ieee_fullname}
\bibliography{final_version}
}

\vspace{-13 mm}
\begin{IEEEbiography}[{\includegraphics[width=1in,height=1.25in,clip,keepaspectratio]{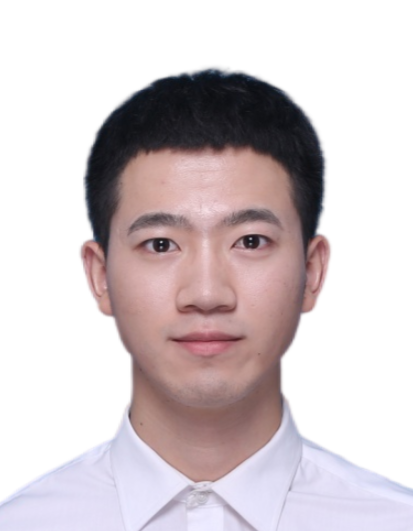}}]
{Ti Wang} is currently a Ph.D. student in the School of Engineeing at École Polytechnique Fédérale de Lausanne (EPFL), Switzerland. He received his Master’s degree from the School of Electronic and Computer Engineering at Peking University (PKU), China, under the supervision of Prof. H. Liu. His research includes 3D pose and shape estimation, 3D computer vision, and behavior understanding.
\end{IEEEbiography}

\begin{IEEEbiography}[{\includegraphics[width=1in,height=1.25in,clip,keepaspectratio]{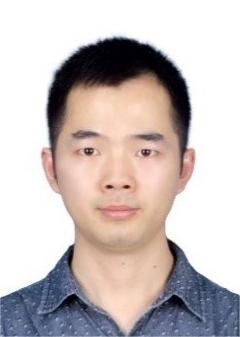}}]{Mengyuan Liu} is an Assistant Professor at Peking University. He received a Ph.D. degree in 2017 under the supervision of Prof. H. Liu from the School of EE\&CS, Peking University (PKU), China. His research interests include human action recognition, human motion prediction, and human motion generation using RGB, depth, and skeleton data. Related methods have been published in T-IP, T-CSVT, T- MM, PR, CVPR, ECCV, ACM MM, AAAI, and IJCAI. He has been invited to be a Technical Program Committee (TPC) member for ACPR, ACM MM, and AAAI.
\end{IEEEbiography}

\begin{IEEEbiography}[{\includegraphics[width=1in,height=1.25in,clip,keepaspectratio]{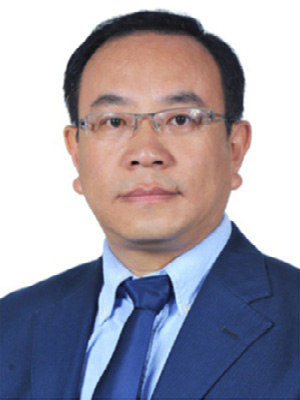}}]{Hong Liu} serves as a Full Professor in the School of EE\&CS, Peking University (PKU), China. He received the Ph.D. degree in Mechanical Electronics and Automation in 1996. Prof. Liu has been selected as Chinese Innovation Leading Talent supported by “National High-level Talents Special Support Plan” since 2013. His research interests include computer vision, pattern recognition and robot motion planning. He is also reviewers for many international journals such as Pattern Recognition, IEEE Trans. on Signal Processing, and IEEE Trans. on PAMI.
\end{IEEEbiography}

\begin{IEEEbiography}[{\includegraphics[width=1in,height=1.25in,clip,keepaspectratio]{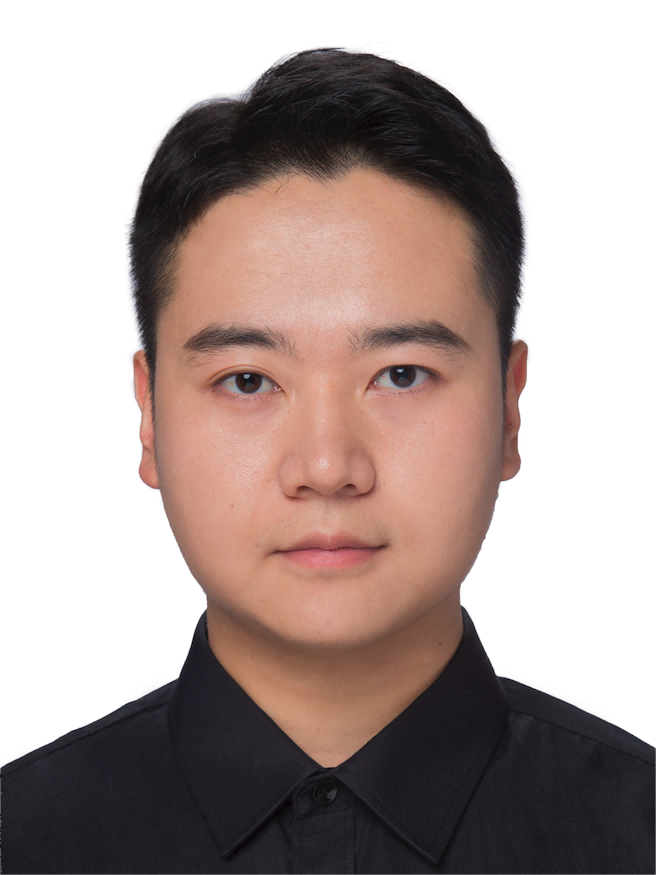}}]{Bin Ren} is a Ph.D. student in the Italian National Artificial Intelligence program co-organized by the University of Pisa and the University of Trento, Italy. He received his Master's degree in 2021 at the School of Electronics and Computer Engineering, Peking University, China. He received his B.Eng. degree in 2016 at the College of Mechanical and Electrical Engineering, Central South University, China.  His research interests lie in explore machine learning to low-level vision and multi-modal LLMs.
\end{IEEEbiography}


\begin{IEEEbiography}[{\includegraphics[width=1in,height=1.25in,clip,keepaspectratio]{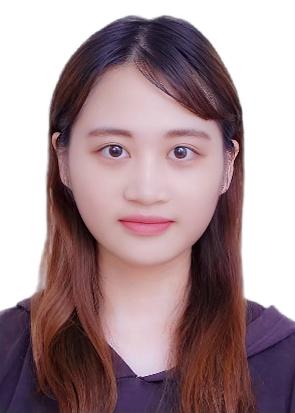}}]
{Yingxuan You} is currently a Ph.D. student in the School of Computer and Communication Sciences at École Polytechnique Fédérale de Lausanne (EPFL), Switzerland. She received her Master's degree from the School of Electronic and Computer Engineering at Peking University (PKU), China, under the supervision of Prof. H. Liu. Her primary research interests include 3D human pose and shape estimation, 3D computer vision, and generative models.
\end{IEEEbiography}

\begin{IEEEbiography}[{\includegraphics[width=1in,height=1.25in,clip,keepaspectratio]{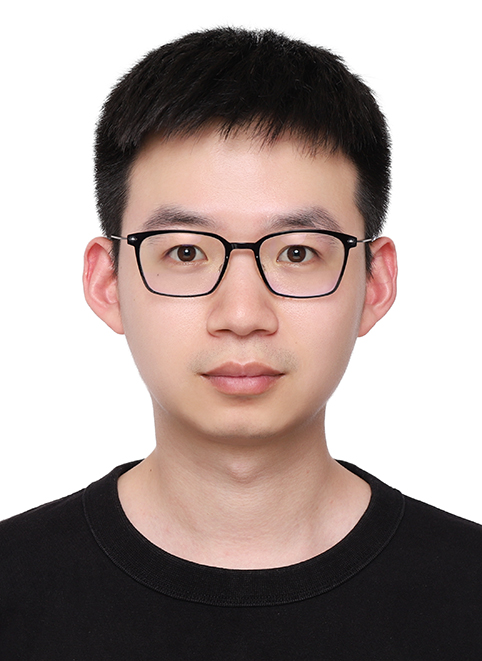}}]{Wenhao Li} is currently a Postdoctoral Researcher at the School of Computer Science and Engineering, Nanyang Technological University, Singapore.  
He received the Ph. D. degree from the School of Computer Science, Peking University, China. 
His research interests lie in deep learning, machine learning, and their applications to computer vision. 
\end{IEEEbiography}

\begin{IEEEbiography}[{\includegraphics[width=1in,height=1.25in,clip,keepaspectratio]{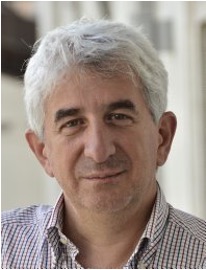}}]{Nicu Sebe} is a Professor at the University of Trento, Italy, where he is leading the research in the areas of multimedia analysis and human behavior understanding. He was the General Co-Chair of the IEEE FG 2008 and ACM Multimedia 2013. He was a program chair of ACM Multimedia 2011 and 2007, ECCV 2016, ICCV 2017 and ICPR 2020. He is a general chair of ACM Multimedia 2022. He is a fellow of IAPR.
\end{IEEEbiography}

\vspace{10 mm}
\begin{IEEEbiography}[{\includegraphics[width=1in,height=1.25in,clip,keepaspectratio]{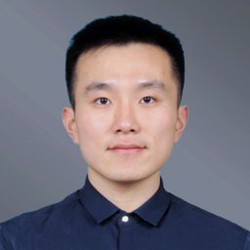}}]{Xia Li} received the B.Eng. degree in computer science from the Beijing University of Posts and Telecommunications in 2017 and the M.Sc. degree in computer science from Peking University in 2020. He received the Ph.D. degree in computer science from ETH Zurich in 2025, where his doctoral studies were part of a joint program with the PSI Center for Proton Therapy.
He is currently a Postdoctoral Researcher with the Department of Information Technology and Electrical Engineering at ETH Zurich. His research interests include medical image processing and machine learning, with a focus on deformable image registration, reconstruction, and segmentation.
\end{IEEEbiography}

 


\end{document}